\documentclass[runningheads]{llncs}

\usepackage{eccv}
\usepackage{eccvabbrv}
\usepackage{graphicx}
\usepackage{booktabs}
\usepackage{algorithm}
\usepackage{algpseudocode}
\usepackage[accsupp]{axessibility}
\usepackage{hyperref}
\usepackage{orcidlink}

\begin{document}

\title{Variational Randomized Smoothing for Sample-Wise Adversarial Robustness}

\author{Ryo Hase\inst{1} \and
Ye Wang\inst{2} \and
Toshiaki Koike-Akino\inst{2} \and
Jing Liu\inst{2} \and
Kieran Parsons\inst{2}}

\authorrunning{R.~Hase et al.}

\institute{Mitsubishi Electric Corporation, Kamakura City, Japan \\
\email{Hase.Ryo@dc.MitsubishiElectric.co.jp}
\and
Mitsubishi Electric Research Laboratories, Cambridge, MA, USA \\
\email{\{yewang,koike,jiliu,parsons\}@merl.com}}

\maketitle

\begin{abstract}
Randomized smoothing is a defensive technique to achieve enhanced robustness against adversarial examples which are small input perturbations that degrade the performance of neural network models. Conventional randomized smoothing adds random noise with a fixed noise level for every input sample to smooth out adversarial perturbations. This paper proposes a new variational framework that uses a per-sample noise level suitable for each input by introducing a noise level selector. Our experimental results demonstrate enhancement of empirical robustness against adversarial attacks. We also provide and analyze the certified robustness for our sample-wise smoothing method.
\end{abstract}

\keywords{Adversarial robustness \and Adversarial examples \and Randomized smoothing \and Median smoothing \and Stochastic regularization \and Meta learning}

\section{Introduction}
\subsection{Background}
Neural networks are vulnerable to adversarial attacks that degrade the performance~\cite{goodfellow-explaining-and-harnessing-2015, szegedy-intriguing-properties-neural-network-2014}. 
Adversarial attacks using adversarial examples can often seriously deteriorate prediction results of a neural network by adding small perturbations to input of the network. 
For example, the Projected Gradient Descent (PGD) attack~\cite{madry2018towards} is widely used example of a white-box adversarial attack that utilizes knowledge of the target neural network such as weights and gradients.
Considering defensive methods for neural networks is crucial to improve robustness against adversarial attacks.

\subsection{Related work}
Various types of defense mechanisms have been proposed to protect neural networks from adversarial attacks.
Adversarial training~\cite{goodfellow-explaining-and-harnessing-2015} enhances the robustness of neural networks by adding adversarial examples as training data. 
For instance, PGD is commonly used to generate the examples used in adversarial training. 
Adversarial purification~\cite{nie2022DiffPure} is a technique to reduce effects of adversarial examples by removing perturbations before they are input to the network. 
Techniques to detect adversarial examples have also been introduced~\cite{Carlini2017adversarial}.

Randomized smoothing\cite{pmlr-v97-cohen19c} is another defensive technique against adversarial examples, the introduces the concept of a {\em smoothed classifier} that counteracts the perturbation of the adversarial examples. 
The process of randomized smoothing perturbs the input with multiple samples of Gaussian noise and aggregates the corresponding outputs of the classifier.
It provides a theoretical certification of robustness that guarantees that the smoothed classifier predicts the correct class, even in the presence of any adversarial perturbations within a certain bound. 

Conventional randomized smoothing perturbs all inputs with the same noise level.
Instead of using the same noise level, Wang {\it et al.}~\cite{wang2020pretrain} introduced an approach to assign sample-wise noise levels to input. 
S{\'u}ken{\'\i}k {\it et al.}~\cite{sukenik2021intriguing} present a theoretical framework to determine sample-wise noise levels. 
Although their framework explains problems of the sample-wise noise level selection, applicability of their approach is limited since it does not work well for relatively larger noise levels.

\subsection{Contributions}
Inspired by the related studies, we propose {\em variational randomized smoothing}, which is a framework utilizing a selector to produce a noise level for every input of a smoothed classifier.
\begin{itemize}
    \item We introduce a {\em variational framework} to build a noise level selector composed of a neural network to determine sample-wise noise levels for randomized smoothing. 
    \item We propose a {\em universal training} scheme using {\em stochastic regularization}, which makes a selector learn various conditions to produce different noise strength at once by randomly sampling regularization parameter $\lambda$.
    \item We improve the controllability in the universal training by using {\em conditional meta learning}, which enables the user to freely adjust noise strength by specifying $\lambda$ at test time without retraining.
    \item Since the selector itself is a neural network which could be potentially a target of adversarial attacks, we also propose a defensive method called {\em dual smoothing} to protect our selector as well as base classifier. 
    \item We provide a modified certified robustness for sample-wise smoothing methods, based on the bound of {\em median smoothing}~\cite{chiang2020detection}. 
    \item Experimental results demonstrate that our proposed methods offer better empirical robustness compared to the conventional randomized smoothing.
\end{itemize}

\section{Randomized smoothing}
\subsection{Fundamental mathematical formulations}

Randomized smoothing\cite{pmlr-v97-cohen19c} is a defense method applied to a base classifier $f: \mathcal{X} \rightarrow \mathcal{C}$, where $\mathcal{X} \subseteq \mathbb{R}^d$ is the input (e.g., image) space and $\mathcal{C} = \{1,2,...,M\}$ is the set of class labels.
In principle, this defense defines an ideal smoothed classifier $g: \mathcal{X} \rightarrow \mathcal{C}$, by choosing the most likely class output
of $f$, when the input is perturbed by Gaussian noise,
\begin{equation}\label{eq_randomized_smoothing_formulation}
g(x) := \underset{c\in \mathcal{C}}{\operatorname{argmax}}\ \mathbb{P} \left[f(x + \varepsilon) = c\right],
\end{equation}
where $\mathbb{P}[\cdot]$ denotes probability with respect to the Gaussian noise $\varepsilon \sim \mathcal{N}(0, \sigma_s^2I_d)$, with $I_d$ denoting the identity matrix of dimensionality $d$.
This technique provides certified robustness\cite{pmlr-v97-cohen19c}, by guaranteeing that the output $g(x + \delta)$ is constant for any adversarial perturbation $\delta \in \mathbb{R}^d$ within $l_2$-radius $R$, i.e., $ \|\delta\|_2 \leq R$, given by
\begin{equation}\label{eq_certification_radius}
    R = \frac{\sigma_s}{2}\left(\Phi^{-1}(p_a) - \Phi^{-1}(p_b) \right),
\end{equation}
where $\Phi^{-1}$ is the inverse of the standard Gaussian Cumulative Distribution Function (CDF), and $p_a$ and $p_b$ are the probabilities of the two most likely outputs of $f(x + \varepsilon)$ for $\varepsilon \sim \mathcal{N}(0, \sigma_s^2I_d)$.

\subsection{Practical randomized smoothing}
\begin{figure}[t]
  \centering
  \begin{subfigure}{0.31\linewidth}    
    \includegraphics[width=\columnwidth]{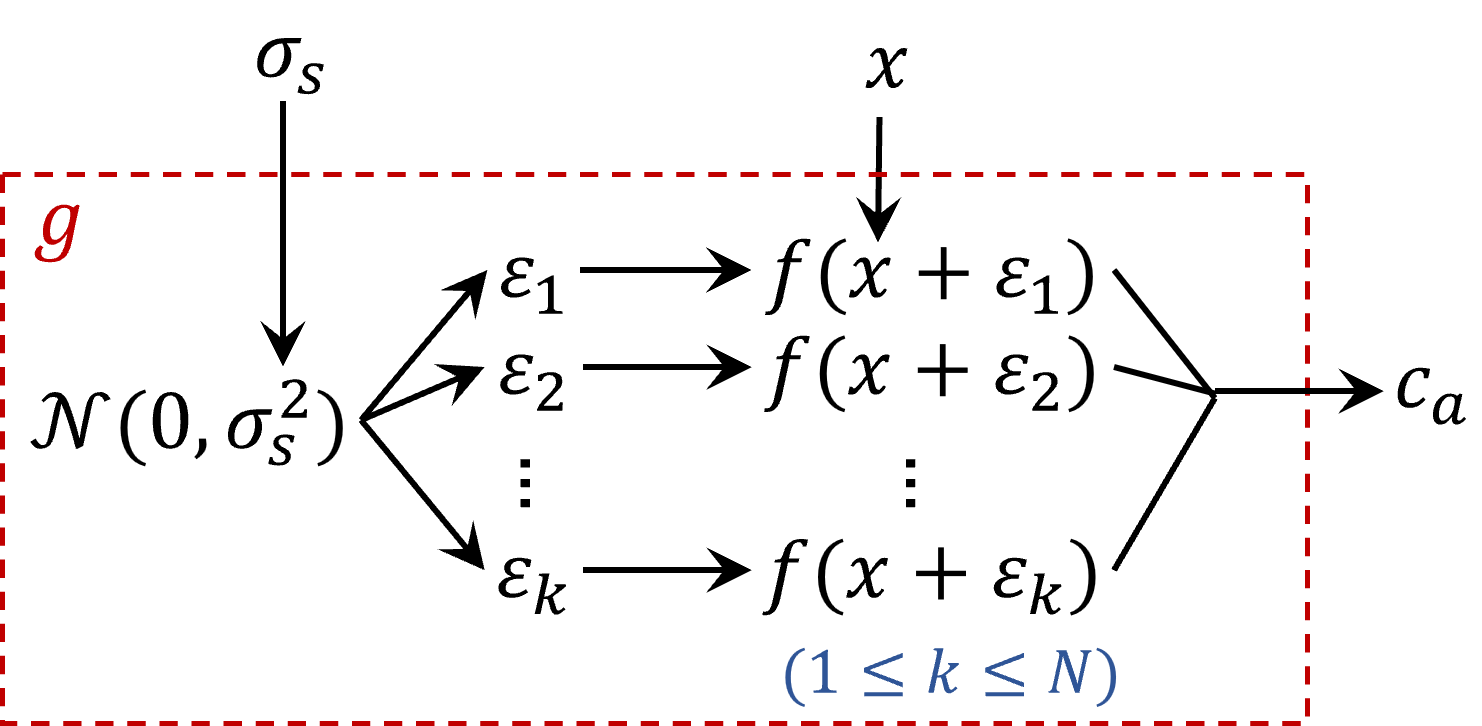}
    \caption{Conventional smoothing $g$.}
    \label{fig_smoothed_classifier_overview}  
  \end{subfigure}
  \hfill
  \begin{subfigure}{0.30\linewidth}
    \includegraphics[width=\columnwidth]{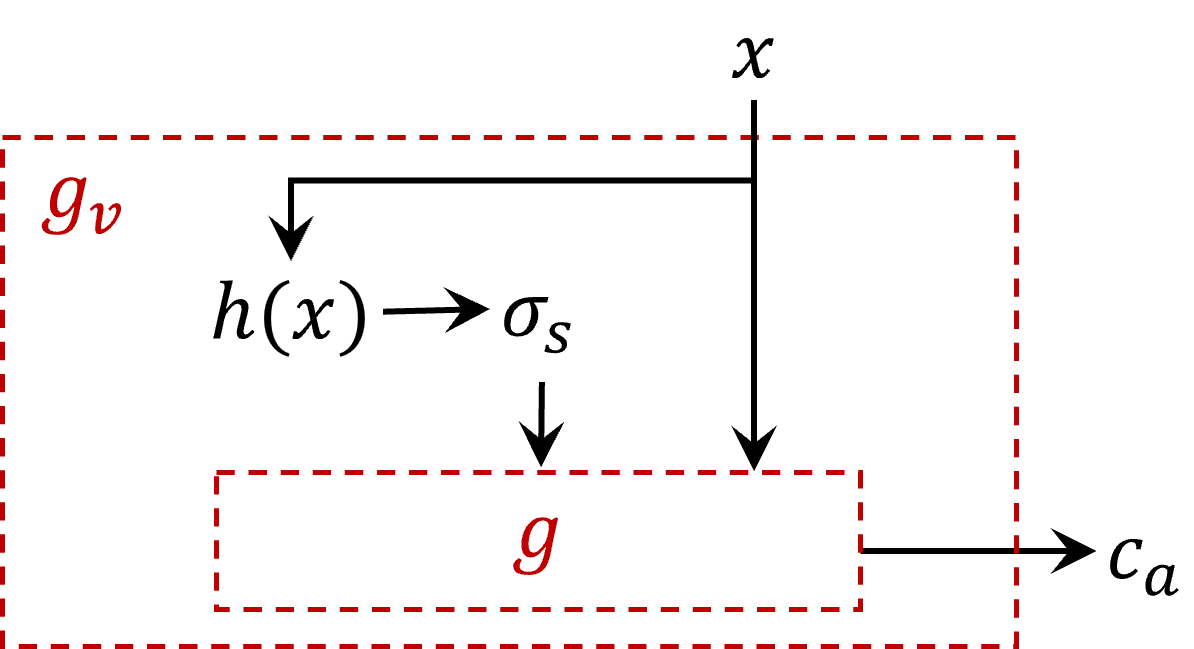}
    \caption{Variational smoothing $g_v$.}
    \label{fig_smoothed_classifier_with_selector_overview}
  \end{subfigure}
  \hfill
  \begin{subfigure}{0.36\linewidth}
    \includegraphics[width=\columnwidth]{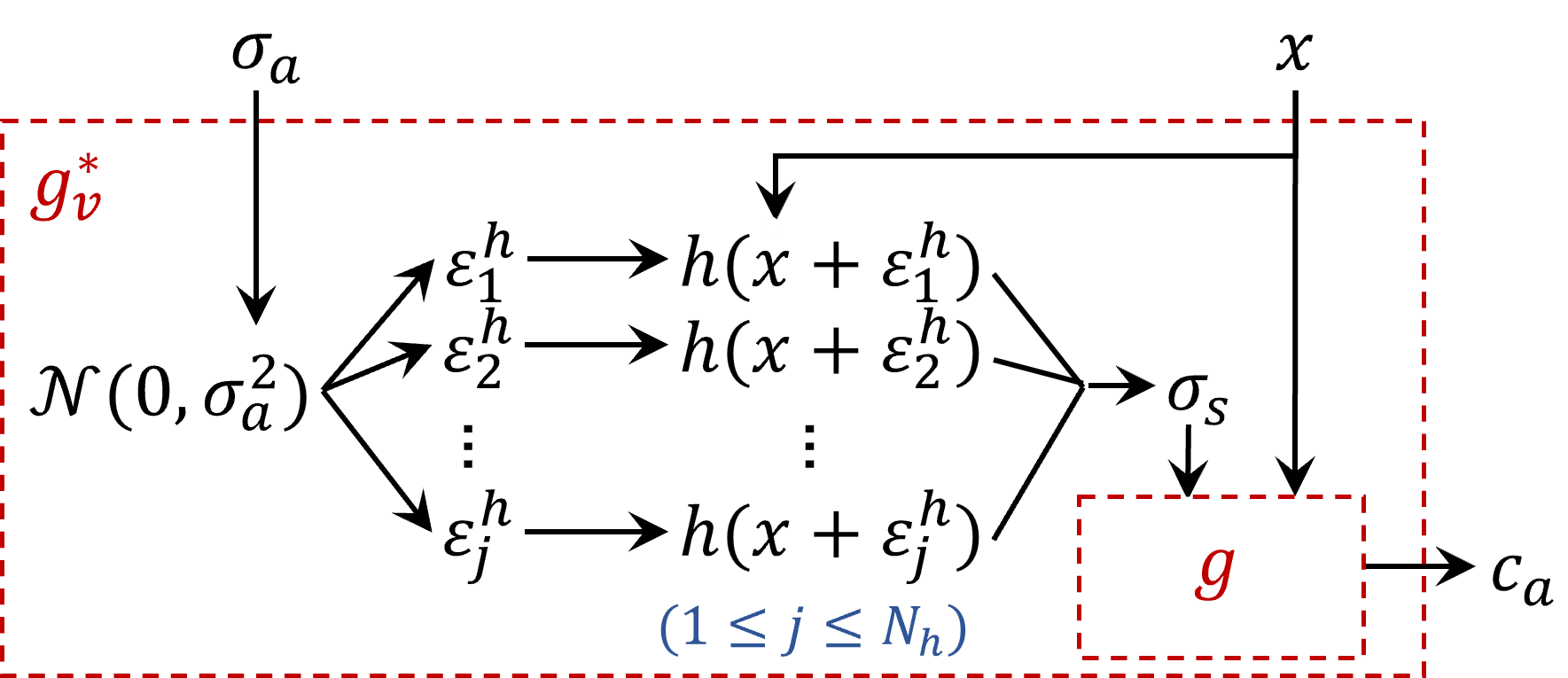}
    \caption{Variational dual smoothing $g_v^*$.}
    \label{fig_smoothed_classifier_with_selector_median_smoothing_overview}
  \end{subfigure}
  \caption{Conventional and proposed approaches of smoothed classifier.}
  \label{fig_smoothed_classifier_overviews}
\end{figure}

As the calculation of the ideal smoothed classifier in~\eqref{eq_randomized_smoothing_formulation} is generally intractable, typically a Monte--Carlo approximation is utilized~\cite{pmlr-v97-cohen19c}. 
Fig.~\ref{fig_smoothed_classifier_overview} depicts an overview of this practical randomized smoothing classifier $g$, which approximates~\eqref{eq_randomized_smoothing_formulation} by taking the majority vote over $N$ samples of Gaussian noise, as given by
\begin{equation}
g(x) \approx \underset{c\in \mathcal{C}}{\operatorname{argmax}}\ 
\sum_{k=1}^{N} \mathbb{I}\left[f(x+\varepsilon_k) = c\right],
\label{eq:majority}
\end{equation}
where $\mathbb{I}[\cdot]$ denotes the binary indicator function, and the Gaussian noise samples are denoted by $\varepsilon_k \stackrel{\text{iid}}{\sim} \mathcal{N}(0, \sigma_s^2I_d)$ for $k \in \{1, 2, \ldots, N\}$.
Note that this algorithm is specified to abstain from making a prediction, if statistical confidence is not satisfied during certification, as described in the following.

Based on the ideal certified radius, given in~\eqref{eq_certification_radius}, a practical certified guarantee is provided by estimating bounds on $p_a$ and $p_b$ for a given confidence level $\alpha$, based on statistical tests applied to the outputs $f(x + \varepsilon_k)$ over the $N$ samples of Gaussian noise~\cite{pmlr-v97-cohen19c}.
Given a confident lower bound $\underline{p_a}$ on the probability $p_a$, we also have an upper bound on $p_b \le 1 - \underline{p_a}$.
Thus, we can confidently certify the radius approximation given by
\begin{equation} \label{eq_certification_radius_practical}
    R \gtrapprox \frac{\sigma_s}{2}\left(\Phi^{-1}(\underline{p_a}) - \Phi^{-1}(1-\underline{p_a}) \right)
    = \sigma_s\Phi^{-1}(\underline{p_a}).
\end{equation}

\subsection{Impact of noise level selection}\label{subsec_impact_noise_level_selection}
An effective and common technique to enhance the performance of randomized smoothing is to train the base classifier $f$ with Gaussian noise augmentation, in order to adapt to the Gaussian noise employed in this defense.
We use $\sigma_a$ to denote the standard deviation of Gaussian noise used for training augmentation.
Hence, with this augmentation, randomized smoothing involves two noise level parameters, $\sigma_s$ and $\sigma_a$. 

\begin{figure}[t]
  \centering
  \begin{subfigure}{0.45\linewidth}    
    \includegraphics[width=\columnwidth]{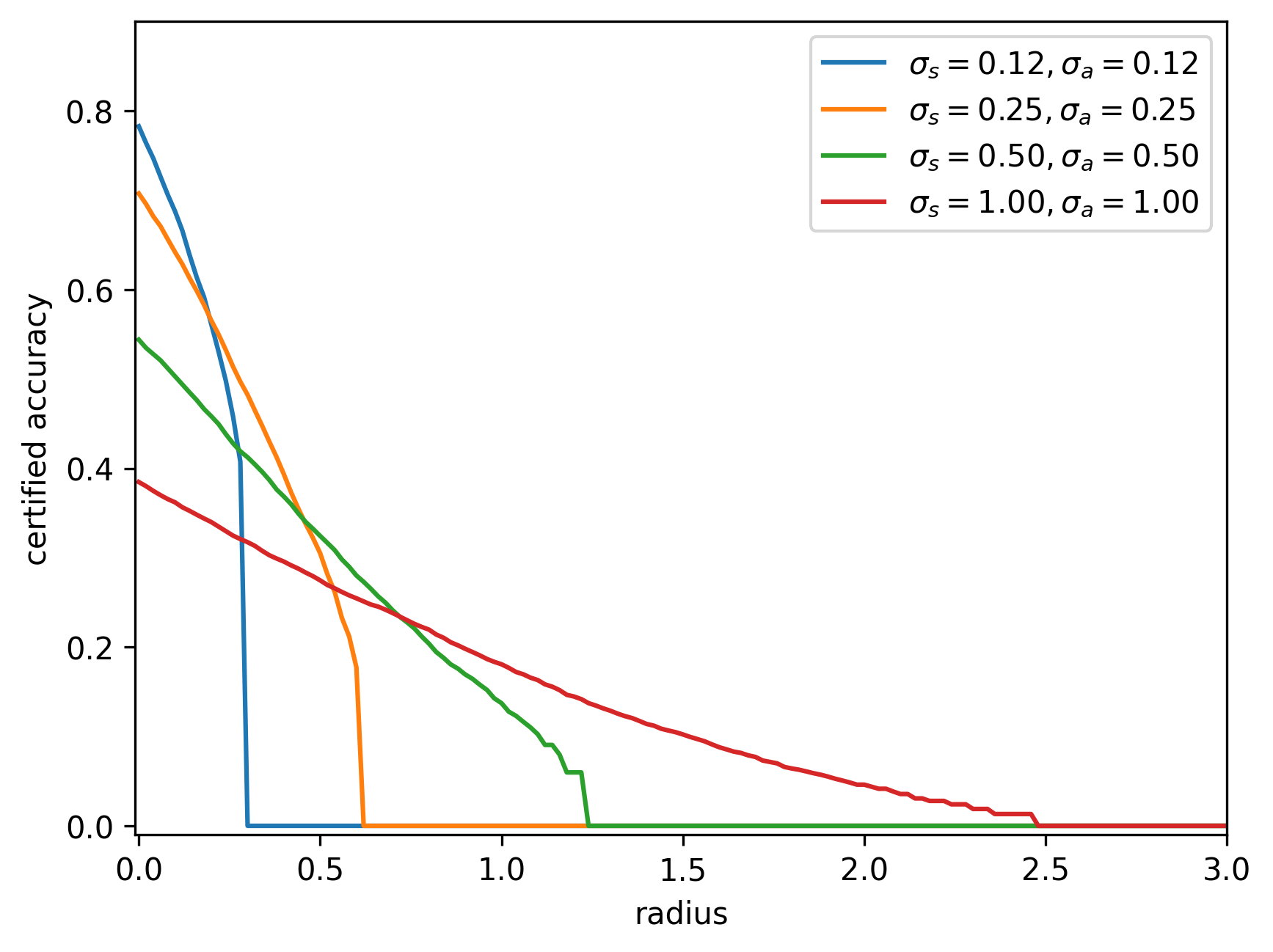}
    \caption{Certified accuracy and radius.}
    \label{fig_cert_acc_randomized_smoothing_example}  
  \end{subfigure}
  \hfill
  \begin{subfigure}{0.45\linewidth}
\includegraphics[width=\columnwidth]{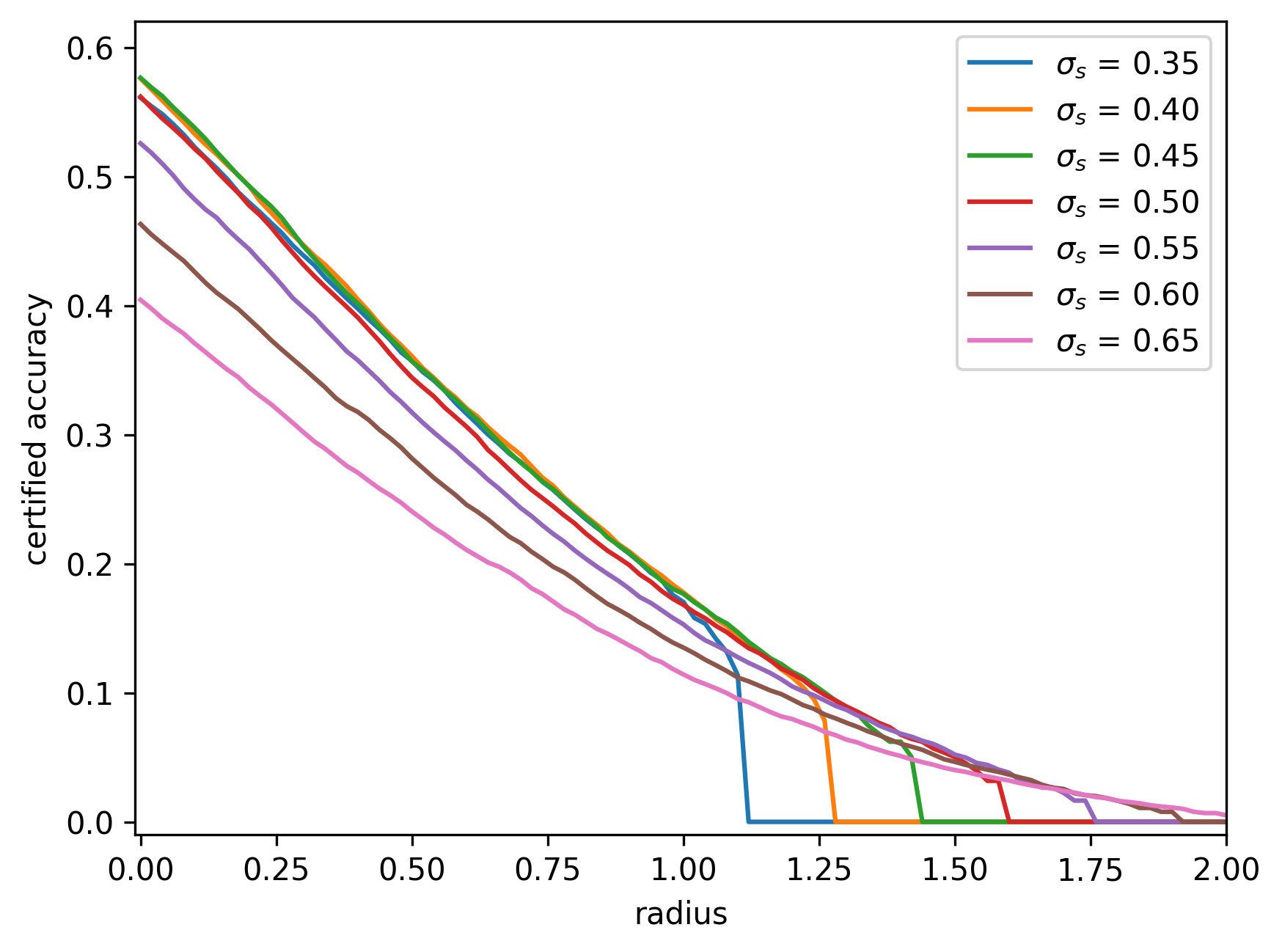}
    \caption{Various $\sigma_s$ at $\sigma_a=0.5$.}
    \label{fig_cert_acc_sigma_a-0.5_varying_sigma_s}
  \end{subfigure}
  \caption{Examples of certified accuracy and radius obtained by randomized smoothing.}
  \label{fig_acc_varying_sigma_s}
\end{figure}

The noise levels $\sigma_s$ and $\sigma_a$ impact the performances of certified accuracy and radius. 
In particular, their selection yields a trade-off, and thus it is often difficult to maximize both certified accuracy and radius together. 
For example, Fig.~\ref{fig_cert_acc_randomized_smoothing_example} shows the certified accuracy obtained by randomized smoothing for the case of $\sigma_s= \sigma_a$.
Here, we trained a base classifier (composed of 4 convolutional layers) on the CIFAR-10 dataset for $400$ epochs, and conducted randomized smoothing with $N=1{,}000$ samples. 
The smaller noise level gives higher certified accuracy, while sacrificing certified radius. 
Fig.~\ref{fig_cert_acc_sigma_a-0.5_varying_sigma_s} shows curves obtained from randomized smoothing for the case of $\sigma_a = 0.5$ while sweeping $\sigma_s$. 
This example also indicates a trade-off between certified accuracy and radius, for various $\sigma_s$ applied to $f$ trained with a fixed $\sigma_a$.

The reason for the trade-off might be explained by the relationship between prediction accuracy and noise level $\sigma_s$. 
It is expected that the classifier would have higher accuracy for smaller $\sigma_s$, which corresponds to increasing the value of $\Phi^{-1}(\underline{p_a})$ in~\eqref{eq_certification_radius_practical}. 
However, the certified radius given by~\eqref{eq_certification_radius_practical} is also proportional to $\sigma_s$.
Hence, realizing the optimal $R$ requires a balance between these values, and the ideal selection of the noise levels $\sigma_s$ and $\sigma_a$ is intractable.
We address this by introducing variational randomized smoothing and universal training methods.

\section{Proposed framework}
\subsection{Noise level selector}
\subsubsection{Basic concept:}
This paper proposes a new {\em variational randomized smoothing} technique to select a suitable $\sigma_s$ for each input image. 
Fig.~\ref{fig_smoothed_classifier_with_selector_overview} shows a high-level overview of this approach.
We use an additional neural network $h: \mathcal{X}\rightarrow [0,\infty)$ to select the randomize smoothing noise level as a function of each input image $x$, i.e., $\sigma_s = h(x)$.
We use $g_v$ to denote the smoothed classifier employing the noise level selector $h$.

\subsubsection{Training formulation:}

The majority voting of the smoothed classifier in~\eqref{eq:majority} is not differentiable, which prevents training $h$.
Thus, for training purposes, we instead use a soft smoothed classifier $g_s$ that aggregates the soft outputs of the model, as given by
\begin{equation}\label{eq_smoother_definition}
    g_s(x) :=
    \frac{1}{N_f^\mathrm{tr}}\sum_{k=1}^{N_f^\mathrm{tr}} \text{softmax}\left( \frac{f_s(x + \varepsilon_k)}{\tau}\right),
\end{equation}
where $f_s$ denotes the soft (logit vector) output of the base classifier $f$,
$\varepsilon_k \stackrel{\text{iid}}{\sim} \mathcal{N}(0, {\sigma_s}^2I_d)$ are $N_f^\mathrm{tr}$ samples of Gaussian noise with $\sigma_s = h(x)$,
and $\tau > 0$ is the temperature parameter for the tempered softmax operation. 
We generally set $\tau=1$ for simplicity.
Note that as $\tau \rightarrow 0$, the soft smoothing is equivalent to the standard majority voting used in~\eqref{eq:majority}.
To train $h$ to pick $\sigma_s$ as a function of $x$ for better accuracy, we employ the typical objective of minimizing cross entropy (CE) loss, $\mathcal{L}_{\text{CE}}(x, y)= -\log g_s(x)[y]$, where $y$ denotes the correct class label for $x$ and $g_s(x)[y]$ denotes the corresponding class likelihood output by $g_s$.
However, minimizing only the CE loss might result in degraded robustness against adversarial attacks as it encourages smaller $\sigma_s$. 

\subsubsection{Stochastic regularization:}

To maintain a reasonable value for $\sigma_s$, we introduce an additional term to regularize $\sigma_s$ towards a desired distribution.
Although the distribution of perturbation $\varepsilon$ is conditionally Gaussian given $\sigma_s$ as $\varepsilon\sim\mathcal{N}(0, \sigma_s^2 I_d)$, it may be no longer marginally Gaussian as $\sigma_s=h(x)$ changes for different inputs $x$.
To encourage Gaussianity of the marginal distribution, we employ a variational framework based on the KL divergence to control the distribution of $\sigma_s$.
Setting the target Gaussian distribution for $\varepsilon$ to be $q = \mathcal{N}(0, \sigma_t^2I_d)$, which captures a target noise level of $\sigma_t$, the KL divergence $D_{\text{KL}}(p\|q)$ to regulate the distribution $p = \mathcal{N}(0, \sigma_s^2I_d)$ is given by
\begin{equation} \label{eq_KL_divergence_term}
\begin{split}
D_{\text{KL}}(p\|q) & = d \bigg[ \frac{1}{2}\left(\frac{\sigma_s}{\sigma_t}\right)^2 - \frac{1}{2} - \log\left(\frac{\sigma_s}{\sigma_t}\right) \bigg]. 
\end{split}
\end{equation}

We use a regularized loss function combining the CE loss and KL divergence,
\begin{equation}\label{eq_two_loss_functions_using_kl_divergence}
    \mathcal{L} =
    (1-\lambda) \mathcal{L}_{\text{CE}}(x, y) + \lambda D_{\text{KL}}(p\|q),
\end{equation}
where $\lambda \in [0,1]$ is a regularization factor adjusting the contribution of each loss term. 
With smaller $\lambda$, the clean data accuracy may be better, while higher robustness may be achieved for higher values of $\lambda$ that encourage $\sigma_s$ to be closer to the target $\sigma_t$.

While the regularization factor $\lambda$ can control the strength of the first and second loss terms, it is cumbersome for us to select the proper $\lambda$ for training. 
Thus, we propose a \emph{universal $\lambda$ training} scheme using a stochastic regularization, which randomly samples $\lambda \sim \text{Uniform}(0,1)$ for each training batch. 
This stochastic regularization aims to train a single selector model $h$ that can flexibly handle the operating tradeoffs across all $\lambda$.
To further improve this meta learning approach, we propose a conditional extension that adds $\lambda$ as an additional input to $h$, i.e., $\sigma_s=h(x, \lambda)$,
to allow flexible control of the noise level and corresponding tradeoff at test time, without the need to retrain the selector $h$.

\subsubsection{Selector training scheme:}

The training procedure for $h$ for each data batch is summarized in Algorithm~\ref{alg_training_process}.

\begin{algorithm}[h]
\caption{Selector $h$ training process (for each batch)}\label{alg_training_process}
\begin{algorithmic}[1]
\State Randomly sample the regularization factor $\lambda \sim \text{Uniform}(0,1)$.
\State Determine noise level $\sigma_s = h(x, \lambda)$ for each input $x$ in the data batch.
\State Apply the soft smoothed classifier, given by~\eqref{eq_smoother_definition}.
\State Evaluate the loss $\mathcal{L}$ for each input, given by~\eqref{eq_two_loss_functions_using_kl_divergence}.
\State Calculate gradient with respect to the total batch loss and update $h$ to minimize.
\end{algorithmic}
\end{algorithm}

\subsubsection{Model architecture:}
\begin{figure}[t]
\centering
\includegraphics[width=0.9\columnwidth]{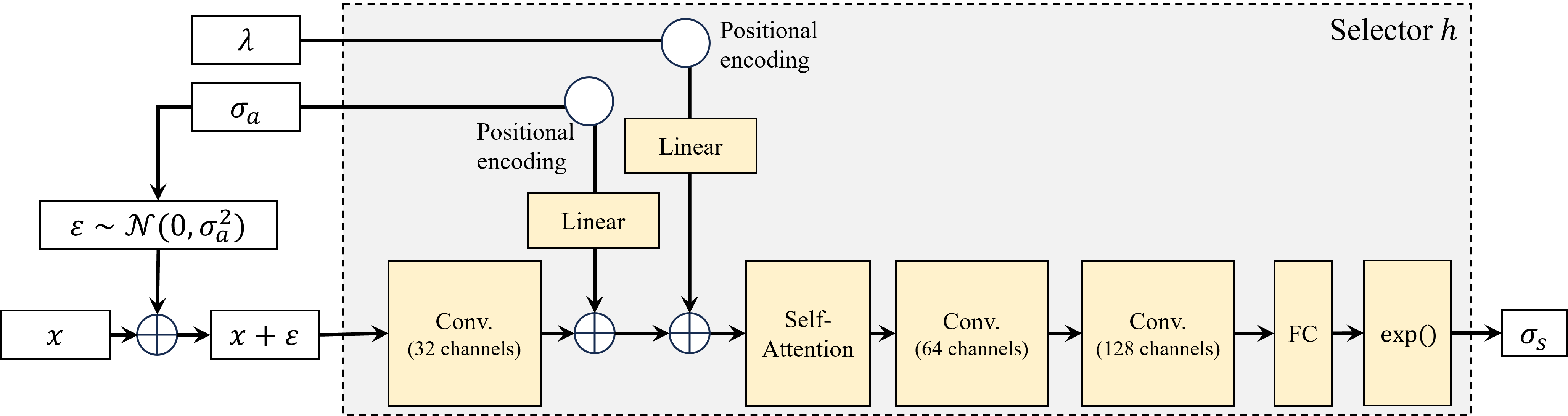}
\caption{Model architecture of noise level selector $h$.}
\label{fig_model_architecture}
\end{figure}

The model architecture of the selector $h$ is depicted in Fig.~\ref{fig_model_architecture}. 
There are three inputs for $h$: $x+\varepsilon$, $\sigma_a$, and $\lambda$. 
The perturbed input $x+\varepsilon$ is directly fed into the first convolutional layer. 
Both $\sigma_a$ and $\lambda$ are used as supporting information for selecting $\sigma_s$. 
Positional encoding and self-attention are used for the inputs $\sigma_a$ and $\lambda$.
The positional encoding layers for $\sigma_a$ and $\lambda$ are added after the first convolution layer, and followed by a self-attention layer.

\subsubsection{Gaussian augmentation in classifier training:}
We train base classifier $f$ with two types of Gaussian augmentation: fixed $\sigma_a$ training and universal $\sigma_a$ training. 
Fixed $\sigma_a$ training is a conventional approach to train $f$ with the same noise level $\sigma_a$ for all input images $x$. 
Training with a fixed $\sigma_a$ is expected to work well with $\sigma_s$ close to $\sigma_a$ as shown in Fig.~\ref{fig_cert_acc_sigma_a-0.5_varying_sigma_s}.
However, it is not straightforward to choose a proper $\sigma_a$ at a training time.

To address this, we apply universal $\sigma_a$ training, which randomly samples $\sigma_a\sim \mathrm{Uniform}(0,\sigma_a')$ for each input $x$ in every training batch. 
This augmentation strategy is inspired by the mixed-noise training introduced in~\cite{wang2020pretrain}. 
Universal $\sigma_a$ training adapts $f$ to be suitable for a wide range of $\sigma_a$.
This offers more flexibility than fixed $\sigma_a$, which would require multiple, separate classifiers for different operating points.
We also employ conditional meta learning by inputting $\sigma_a$ as conditional information to the universally trained classifier so that its operation point is adjustable at test time.

\subsection{Enhancement of selector robustness}
\subsubsection{Randomized smoothing for the selector:}
Since $h$ is also a neural network component, it is possible for an adversarial input attack to cause $h$ to select a $\sigma_s$ that performs poorly for randomized smoothing.
Thus, a defense should also be applied to $h$, however, in the previous section, we described the selector $h$ without defense.
The conventional randomized smoothing techniques for classification tasks is not readily applicable, since $h$ selects continuous noise levels rather than discrete class labels.
Hence, we employ median smoothing~\cite{chiang2020detection}, which is an extension of randomized smoothing for regression problems. 

\subsubsection{Median smoothing:} 
Median smoothing uses the median of multiple regressor outputs for Gaussian augmented input as the smoothed prediction result. 
We denote the smoothed classifier, using $h$ with median smoothing, as $g_v^*$, which is illustrated in Fig.~\ref{fig_smoothed_classifier_with_selector_median_smoothing_overview}. 
Similar to the conventional randomized smoothing, the input image $x$ is perturbed with Gaussian noise $\varepsilon \sim \mathcal{N}(0, \sigma_m^2 I_d)$ where $\sigma_m > 0$ is the median smoothing noise level. 
Let $h_p(x + \varepsilon)$ denote the $p$th percentile of the output of $h(x + \varepsilon)$, with respect to the statistics of the Gaussian input perturbation.
Median smoothing uses the median, $\sigma_s = h_{50\%}(x + \varepsilon)$, as the smoothed result of $h$.
In practice, for both selector training and at test time, this median is empirically computed from multiple samples.
This smoothed output $\sigma_s$ is used for successive randomized smoothing of the base classifier $f$.
Thus, $g^*_v$ employs a dual smoothing to protect both $h$ and $f$.

\begin{figure}[tb]
  \centering
  \begin{subfigure}{0.32\linewidth}
    \includegraphics[width=\columnwidth]{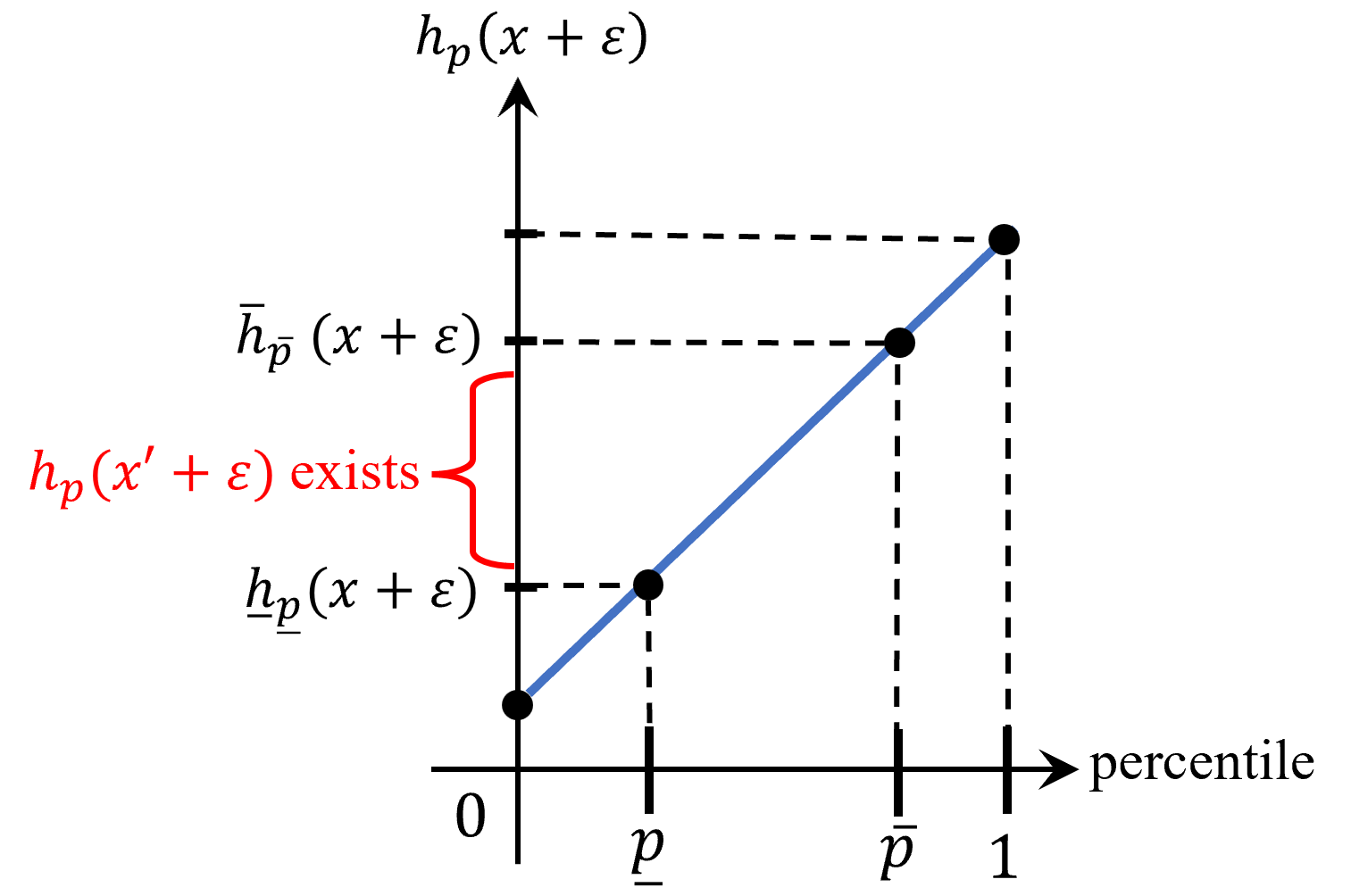}
    \caption{Theoretical bounds.}
    \label{fig_theo_bound_graph}
  \end{subfigure}
  \hfill
  \begin{subfigure}{0.65\linewidth}
    \includegraphics[width=\columnwidth]{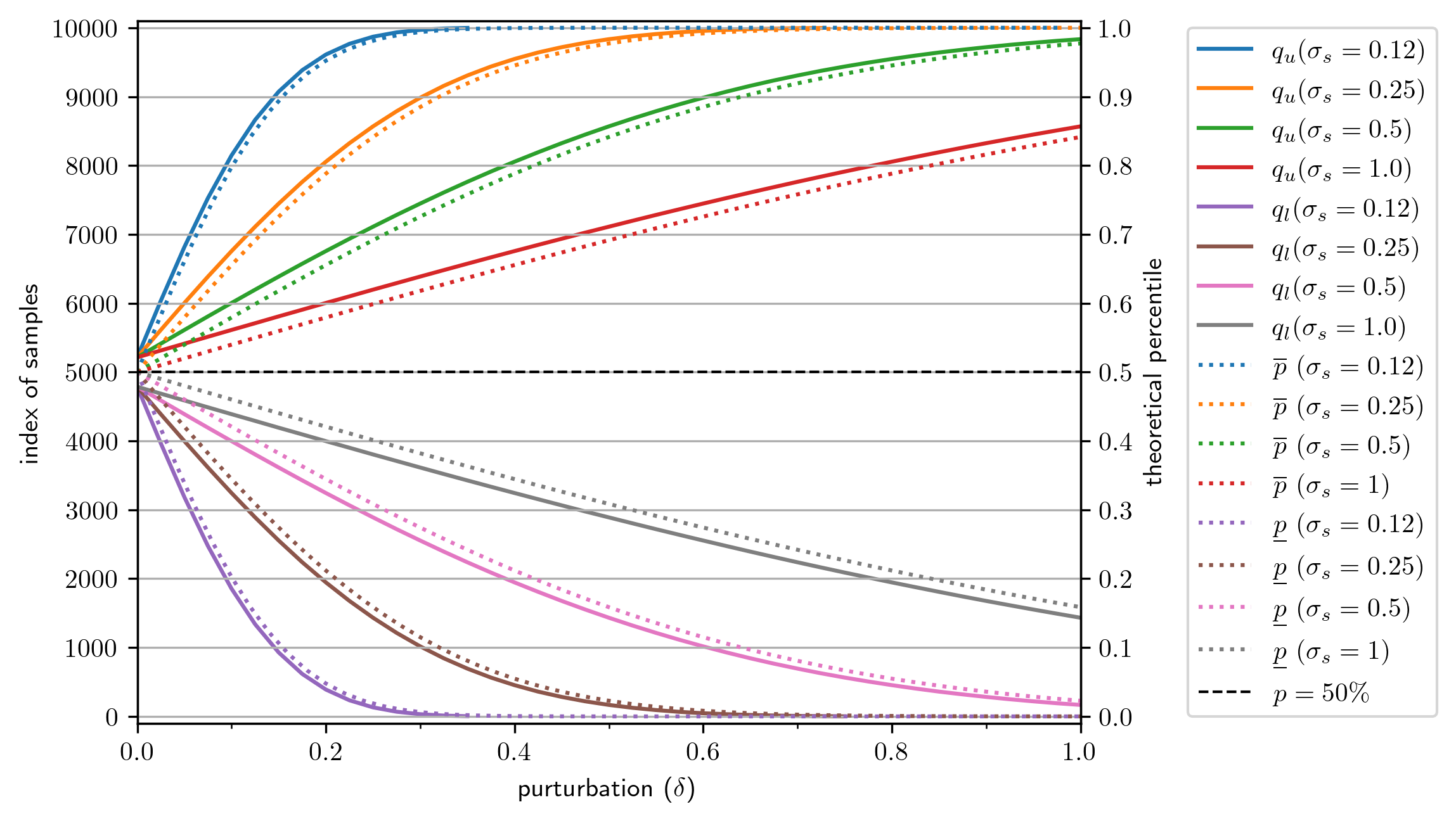}
        \caption{Difference between theoretical and empirical bounds.}
    \label{fig_empirical_bound_graph}
  \end{subfigure}
  \caption{Upper and lower bounds of $h_p(x'+\varepsilon)$.}
  \label{fig:theoretical_empirical_bound_graph}
\end{figure}

Similar to conventional randomized smoothing, median smoothing provides guarantees in the form of upper and lower bounds on the output in the presence of any adversarial perturbation $\delta \in \mathbb{R}^d$, within a given radius ${\|\delta\|}_2 < D$.
We use $\underline{h}$ and $\overline{h}$ to denote lower and upper bounds of output of $h$, respectively, and the shorthand $x' := x + \delta$.
For any perturbation $\delta \in \mathbb{R}^d$, with ${\|\delta\|}_2 < D$, median smoothing guarantees upper and lower bounds on the median smoothed output, given by
\begin{equation}\label{median_smoothing_theoretical_upper_and_lower_bounds}
\underline{h}_{\underline{p}}(x+\varepsilon) \leq h_p(x' + \varepsilon) \leq \overline{h}_{\overline{p}}(x+\varepsilon),
\end{equation}
where $\underline{p} = \Phi(\Phi^{-1}(p)-D/\sigma_m)$ and $\overline{p} = \Phi(\Phi^{-1}(p)+D/\sigma_m)$. For the case of median ($p=50\%$), we have $\underline{p} = \Phi(-D/\sigma_m)$ and $\overline{p} = \Phi(D/\sigma_m)$.
Fig.~\ref{fig_theo_bound_graph} depicts the concept of these bounds.

It is intractable to determine the exact distributions and percentiles for the bounds given above.
Hence,~\cite{chiang2020detection} provides a Monte--Carlo method to approximate the values of $\underline{h}_{\underline{p}}$ and $\overline{h}_{\overline{p}}$.
Their algorithm first generates $N_h$ samples of $h(x + \varepsilon_k)$, for $k \in \{1,2,\ldots,  N_h\}$. 
Then, the samples are sorted based on their magnitude in an ascending order. 
Let $h_q$ denote the value with the sorted index $q$. 
Their algorithm determines indices $q_l$ and $q_u$ that correspond to the lower bound $\underline{p}$ and upper bound $\overline{p}$ within confidence level $\alpha_h$. 
Using $q_l$ and $q_u$, empirical upper and lower bounds of $h(x' + \varepsilon)$ are determined as $\underline{h}_{q_l}(x+\varepsilon)$ and $\overline{h}_{q_u}(x+\varepsilon)$, respectively.
Fig.~\ref{fig_empirical_bound_graph} indicates differences between empirically determined indices $q_u$ and $q_l$ and theoretical percentiles $\overline{p}$ and $\underline{p}$ for $N_h=10{,}000$. 
The gaps between theoretical and empirical percentiles become smaller as $N_h$ increases as shown in Appendix A.

\subsubsection{Certified accuracy and radius for sample-wise smoothing:}
Certified robustness can be discussed using the upper and lower bounds on the output of $h$ determined by median smoothing. 
Let $Q$ be the set of indices satisfying $Q=\{q\in\mathbb{Z} \mid q_l \leq q \leq q_u\}$. 
The number of possible $\sigma_s=h_q(x+\varepsilon)$ is $|Q|$, and the certified accuracy and radius for $x$ is analyzed across all possible $\sigma_s$. 
We take the worst case of certified accuracy and radius across the possible $\sigma_s$ to provide guaranteed lower bounds on certified accuracy and radius. 
If at least one incorrect label or abstained prediction is found in the pairs, the certified radius taken to be $R=0$.
Otherwise, the smallest value of $R$ is used.
Additionally, we set $R=D$ in the case of $R>D$, since $q_u$ and $q_l$ might be different if the perturbation $\|\delta\|_2 >D$.

Since the bounds on the selector output $h(x' + \varepsilon)$ might still allow large deviations, we also propose clipping to prevent larger deviations, i.e., clamping the output to firmly guarantee $h_l \leq h(x'+\varepsilon) \leq h_u$, for some parameters $h_l$ and $h_u$.
In Appendix B, examples of $\underline{h}_{q_l}$ and $\overline{h}_{q_u}$ with a set of $D$ are shown to illustrate the deviations from desired output. 

\section{Experimental results}
\subsection{Overview}
\subsubsection{Types of experiments:}
We conducted experiments to examine the performance of variational randomized smoothing and compare with the conventional approach, which uses fixed $\sigma_s$. 
We consider two types of experiments:
\begin{enumerate}
    \item[A.] Certified robustness: 
    Certified accuracy and radius were analyzed. 
    Specifically, this demonstrates how attacks to both $h$ and $f$ affect certified accuracy and radius of our proposed method.
    \item[B.] Empirical robustness: 
    Classification accuracy was examined with clean images and images perturbed by adversarial attacks. 
    This demonstrates the practical performance of our method, compared to the certified lower bounds.
\end{enumerate}

\subsubsection{Model and dataset.}
The base classifier $f$ is a neural network mainly composed of 4 convolutional layers. 
It was trained on the CIFAR-10 dataset with Gaussian noise augmentation for 400 epochs. 
For fixed $\sigma_a$ training, the augmentation noise $\sigma_a$ is chosen from $0.12$, $0.25$, $0.50$, and $1.00$. 
Additionally, another model $f_u$ is trained with universal $\sigma_a$ training, by randomly sampling $\sigma_a \sim \mathrm{Uniform}(0,1)$ during training, i.e., $\sigma_a' = 1$.

Selector models $h$ are trained for 200 epochs for each base model described above, with the same corresponding data augmentation, and parameters $N_f^\mathrm{tr} = N_h^\mathrm{tr}= 10$. 
For the selectors trained with the fixed $\sigma_a$ base models $f$, the corresponding $\sigma_a$ is used as the input to $h$. 
For the case of selector $h$ trained for the universal $\sigma_a$ model $f_u$, the $\sigma_a$ input of $h$ was set to $0.5$.
We use a target noise level of $\sigma_t = 2 \sigma_a$. 

\subsection{Experiment A: certified accuracy and radius}
\subsubsection{Conditions:}
For the baseline conventional smoothed classifier $g$, we consider the following two cases:
\begin{enumerate}
    \item[a)] $g$ with $\sigma_s$ sweeping: 
    This baseline indicates the maximum possible values of certified accuracy and robustness if $\sigma_s$ can be ideally adjusted for any operating point. 
    We plot the envelope determined by the maximum values in the certified accuracy and radius curves, while sweeping across $\sigma_s$. 
    For example, taking the maximum of the curves in Fig.~\ref{fig_cert_acc_sigma_a-0.5_varying_sigma_s} yields the envelope.
    
    \item[b)] $g$ with $\sigma_s=\sigma_a$: 
    This is the baseline for certified accuracy and radius obtained by using the same noise level for training augmentation and test time smoothing. 
    This is the conventional approach for randomized smoothing.
\end{enumerate}

For the proposed methods, we consider the following two scenarios:
\begin{enumerate}
    \item[i)] $g_v^*$ with attack on $h$: 
    Certified accuracy and radius can be determined based on the process described in the previous section. 
    However, it is practically hard to conduct certification for $\sigma_s=h_q(x+\varepsilon_j)$ with all possible $q\in Q$. 
    Hence, we picked up three $\sigma_s$ selected by $h$: $\underline{h}_{q_l}$, $\overline{h}_{q_u}$, and median $h_{50\%}$. 
    Then, certified accuracy and radius of $g_v^*$ are obtained approximately using these $\sigma_s$. 
    To compare certified accuracy and radius of $g_v^*$ with a baseline envelope, we chose $D$ and $\lambda$ as shown in Table~\ref{table_distortion_budget_in_experiments} and determined an envelope. 
    
    \item[ii)] $g_v^*$ with no attack on $h$: 
    For comparison, we also derived the certified accuracy and radius if there is no attack on $h$, but only on $f$. 
    This shows an idealistic certified robustness, if suitable $\sigma_s$ determined by $h$ for each input can be used in randomized smoothing. 
    When comparing with scenario (i), we can analyze how certified accuracy will degrade by attacks on $h$. 
    By choosing the median of $h(x+\varepsilon)$ as $\sigma_s$, certified accuracy for this weaker attack scenario can be obtained. 
    An envelope of this certified accuracy was determined across multiple $\lambda$ as shown in Table~\ref{table_distortion_budget_in_experiments}.
\end{enumerate}
To determine $q_l$ and $q_u$, we use $N_h=1{,}000$ and confidence level $\alpha_h=10^{-5}$. 
Regarding certification for $f$, we use $N=1{,}000$ and a confidence level $\alpha=10^{-3}$. 
Other parameters $\lambda$, $h_l$, and $h_u$ for each $\sigma_a$ are listed in Table~\ref{table_clipping_threshold}.

\begin{table}[tb]
    \footnotesize
    \begin{tabular}{cc}
      \begin{minipage}[t]{.52\hsize}
        \centering
        \caption{$D$ and $\lambda$ per $\sigma_a$}
        \begin{tabular}{ |c|l|l| } 
         \hline
         $\sigma_a$ & $D$ & $\lambda$  \\ \hline\hline
         0.12 & 0.0, 0.05, 0.1, 0.2 & 0.0, 0.1, 0.2\\ \hline
         0.25 & 0.0, 0.05, 0.1, 0.2, 0.3 & 0.0, 0.1, 0.2\\ \hline
         0.50 & 0.0, 0.05, 0.1, 0.2, 0.3 & 0.0 \\ \hline
         1.00 & 0.0, 0.05, 0.1, 0.2, 0.3, 0.4 & 0.0 \\
         \hline
        \end{tabular}
        \label{table_distortion_budget_in_experiments}
      \end{minipage}
      \begin{minipage}[t]{.45\hsize}
        \centering
        \caption{$h_l$ and $h_u$ per $\sigma_a,\lambda$}
        \begin{tabular}{ |c|l|l|l||c|l|l|l| } 
         \hline
         $\sigma_a$ & $\lambda$& $h_l$ & $h_u$&$\sigma_a$ & $\lambda$& $h_l$ & $h_u$ \\ \hline\hline
         0.12 & 0.0 & 0.06 & 0.10 & 0.25 & 0.1 & 0.18 & 0.25\\ \hline
         0.12 & 0.1 & 0.08 & 0.11 & 0.25 & 0.2 & 0.20 & 0.27\\ \hline
         0.12 & 0.2 & 0.09 & 0.12 & 0.50 & 0.0 & 0.34 & 0.48\\ \hline
         0.25 & 0.0 & 0.16 & 0.24 & 1.00 & 0.0 & 0.68 & 1.10\\ \hline
        \end{tabular}
        \label{table_clipping_threshold}
      \end{minipage}
    \end{tabular}
    \label{table_experiment_parameters}
\end{table}

\subsubsection{Results:}
\begin{figure}[t]
  \centering
  \begin{subfigure}{0.45\linewidth}    
    \includegraphics[width=\columnwidth]{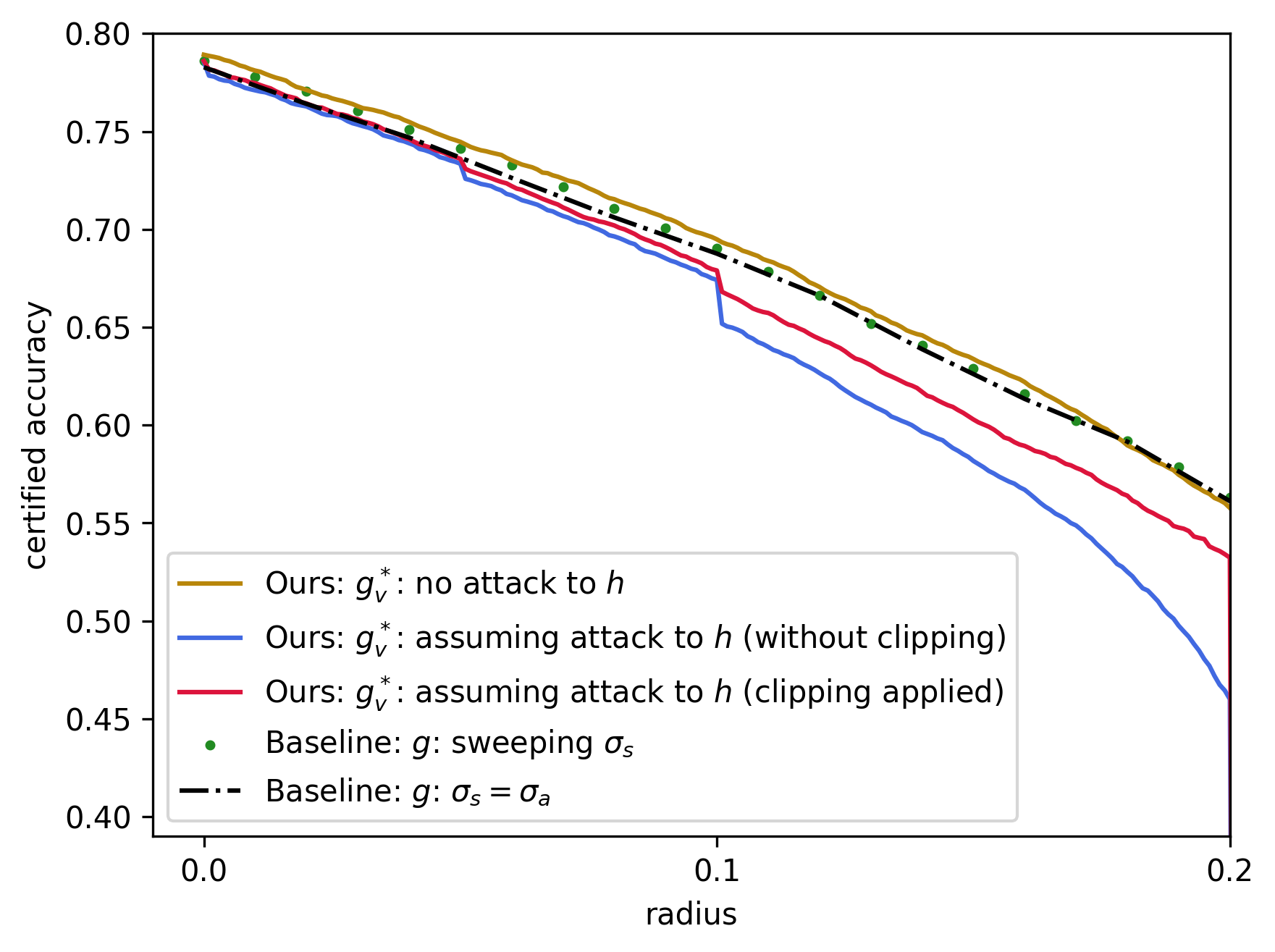}
        \caption{$\sigma_a=0.12$.}
        \label{fig_result_cert_acc_sigma-a0.12}
  \end{subfigure}
  \hfill
  \begin{subfigure}{0.45\linewidth}
\includegraphics[width=\columnwidth]{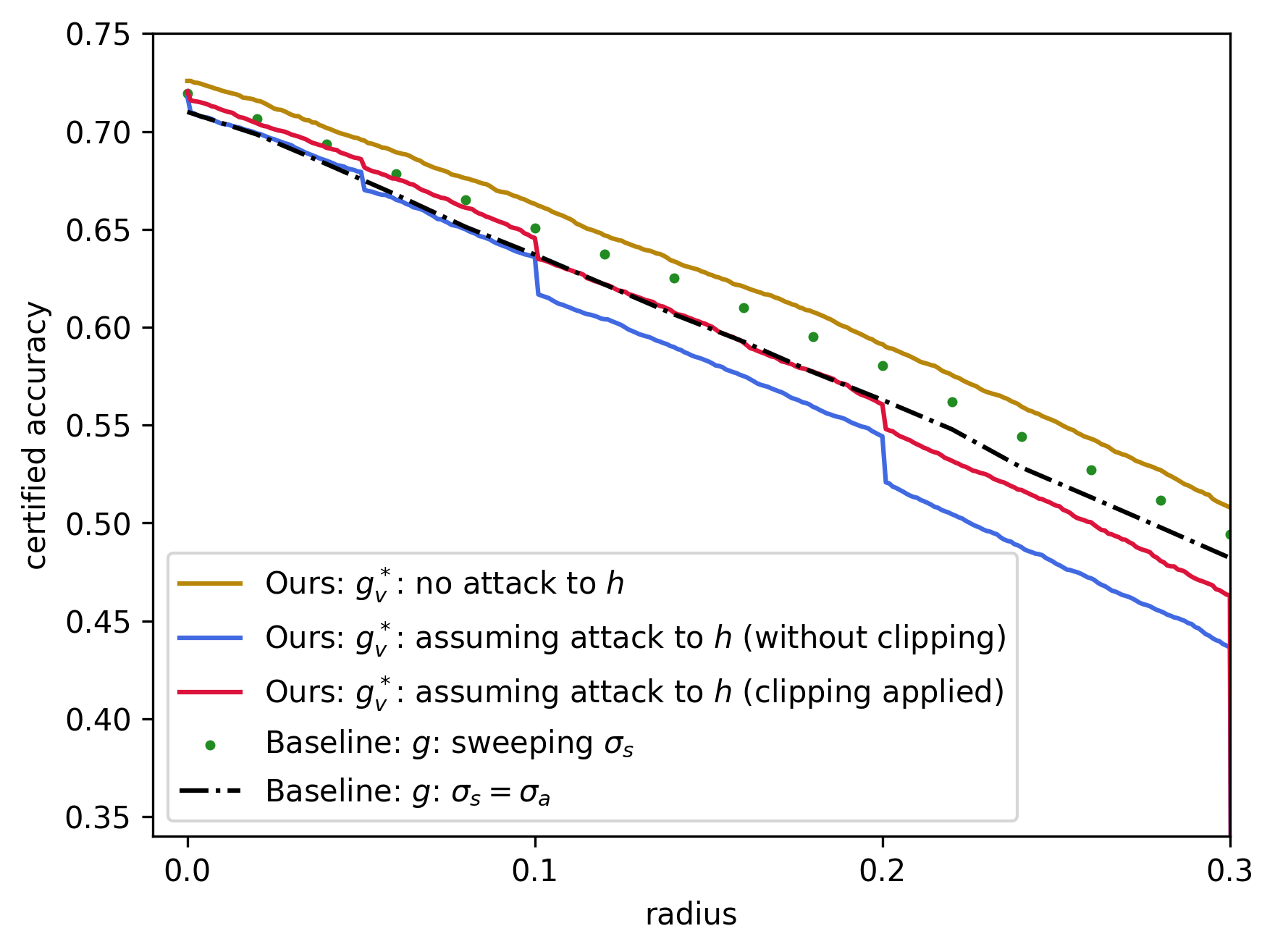}
        \caption{$\sigma_a=0.25$.}
        \label{fig_result_cert_acc_sigma-a0.25}
  \end{subfigure}\\
  \begin{subfigure}{0.45\linewidth}    
    \includegraphics[width=\columnwidth]{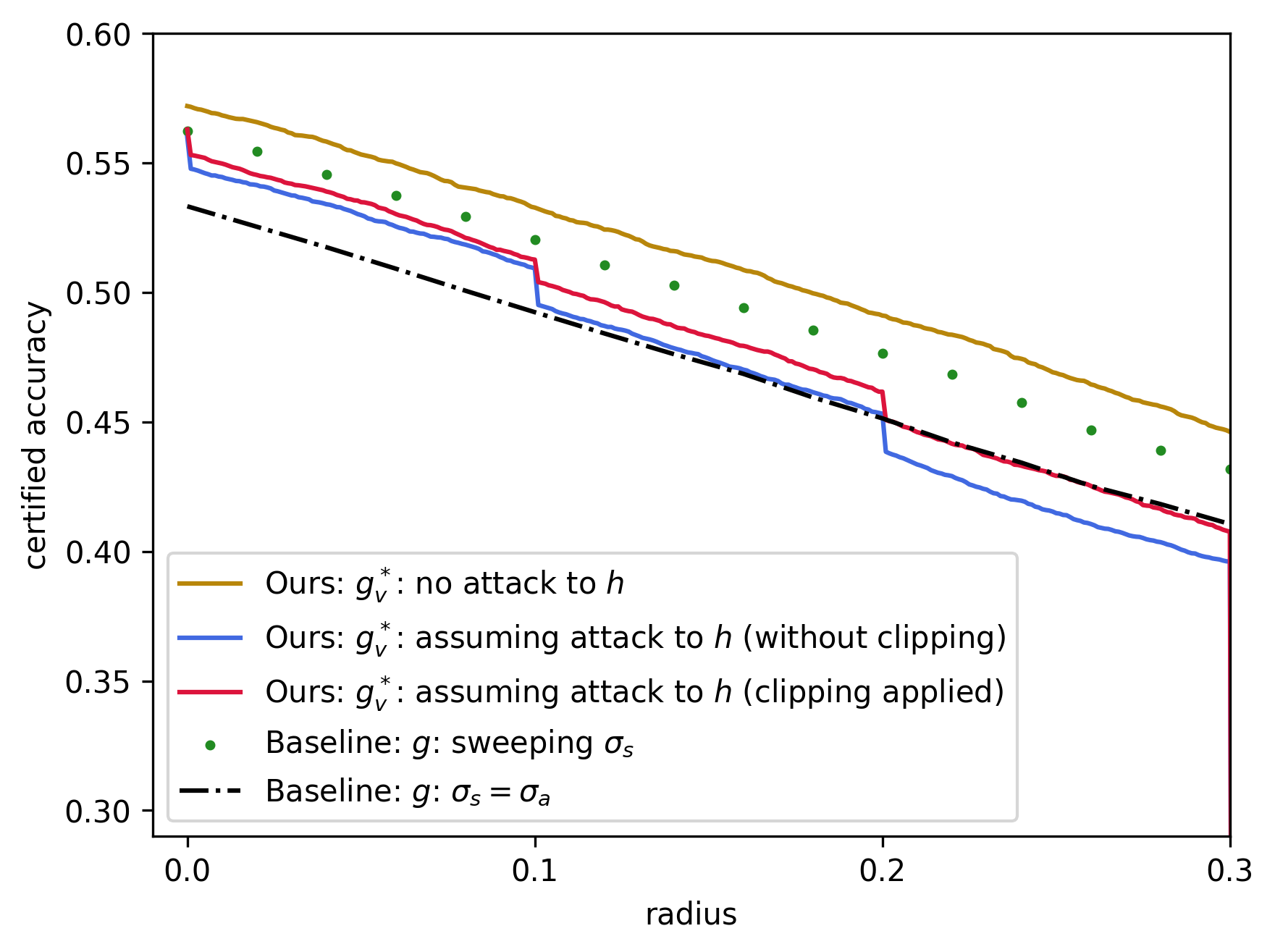}
        \caption{$\sigma_a=0.50$.}
        \label{fig_result_cert_acc_sigma-a0.50}
  \end{subfigure}
  \hfill
  \begin{subfigure}{0.45\linewidth}
\includegraphics[width=\columnwidth]{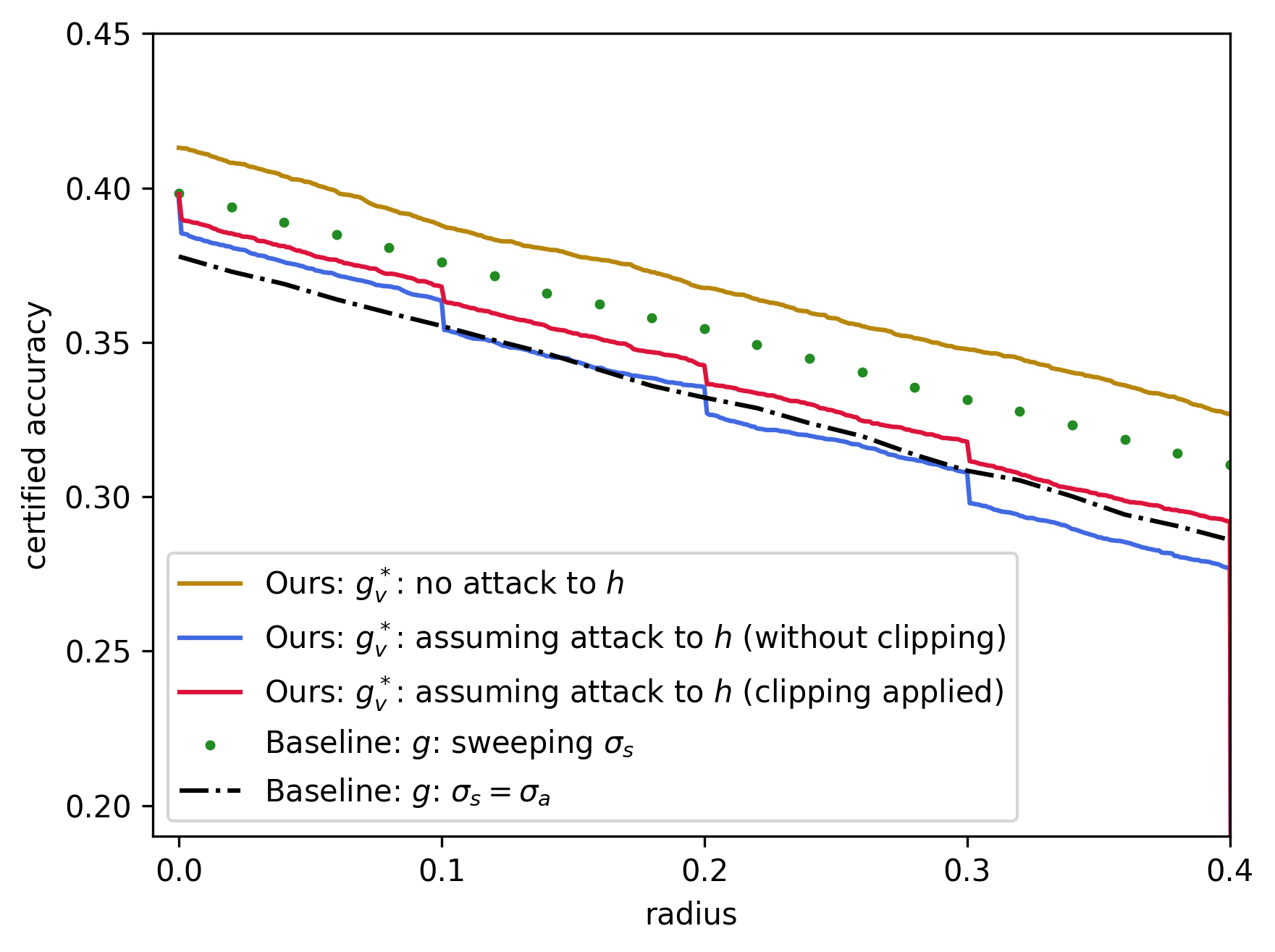}
        \caption{$\sigma_a=1.00$.}
        \label{fig_result_cert_acc_sigma-a1.00}
  \end{subfigure}
  \caption{Certified accuracy and radius with proposed method $g_v^*$ and baseline $g$.}
\label{fig_result_cert_acc_and_radius_experiment_a}
\end{figure}

Fig.~\ref{fig_result_cert_acc_and_radius_experiment_a} shows the results of Experiment A. 
The performance of our method $g_v^*$ with no attack to $h$ was better than that of baselines. 
This suggests that sample-wise smoothing is advantageous if adversary does not attack $h$. 
In the presence of an attack on $h$, the performance was still close to baselines if $D$ was relatively small.
However, the performance can degrade as radius becomes larger. 
We see that applying clipping helps to alleviate the performance degradation of $g_v^*$. 

\subsection{Experiment B: empirical clean and robust accuracy}
\subsubsection{Conditions:}
To evaluate practical adversarial robustness of our method, the commonly used PGD attack is employed to produce adversarial examples. 
We consider two levels of attack strength as follows:
\begin{enumerate}
    \item[1)] Weaker attack: 
    We assume only base classifier $f$ is the attack target. 
    This is a weaker attack setting because PGD attack does not use any information about $g$ and $h$, but only $f$.

    \item[2)] Stronger attack: 
    We assume that the PGD attack can use information of $g$ and $h$ as well as $f$. 
    For $g$, the PGD attack knows $\sigma_s$ for smoothing to discover better adversarial examples. 
    Likewise, the PGD attack has knowledge of the noise level target $\sigma_t$, $\sigma_a$ and $\lambda$ for $h$ in $g_v$ and $g_v^*$. 
    This knowledge enhances the PGD attack strength for smoothed classifiers $g$, $g_v$, and $g_v^*$. 
\end{enumerate}

We consider the tradeoff between clean accuracy and robust accuracy. 
Clean accuracy is the performance on clean data (without any attack), while the randomized smoothing defense is applied. 
For robust accuracy, we evaluate performance with the PGD attack for a specific distortion budget $\gamma$.
Accuracy of the smoothed classifiers is defined by top-ranked class determined by randomized smoothing over $N=1{,}000$ samples. 
Regarding $g_v$ and $g_v^*$ for fixed $\sigma_a$ models, $\lambda$ as a parameter for $h$ is taken from the set $\{0,0.1,0.2,\dots,0.4\}$ to produce each point. 
Further, $g_v^*$ for the universal $\sigma_a$ model uses $\lambda \in \{0,0.05,0.1,0.15,\dots,0.9\}.$

\subsubsection{Weaker attack results:}
\begin{figure}[t]
  \centering
  \begin{subfigure}{0.42\linewidth}    
    \includegraphics[width=\columnwidth]{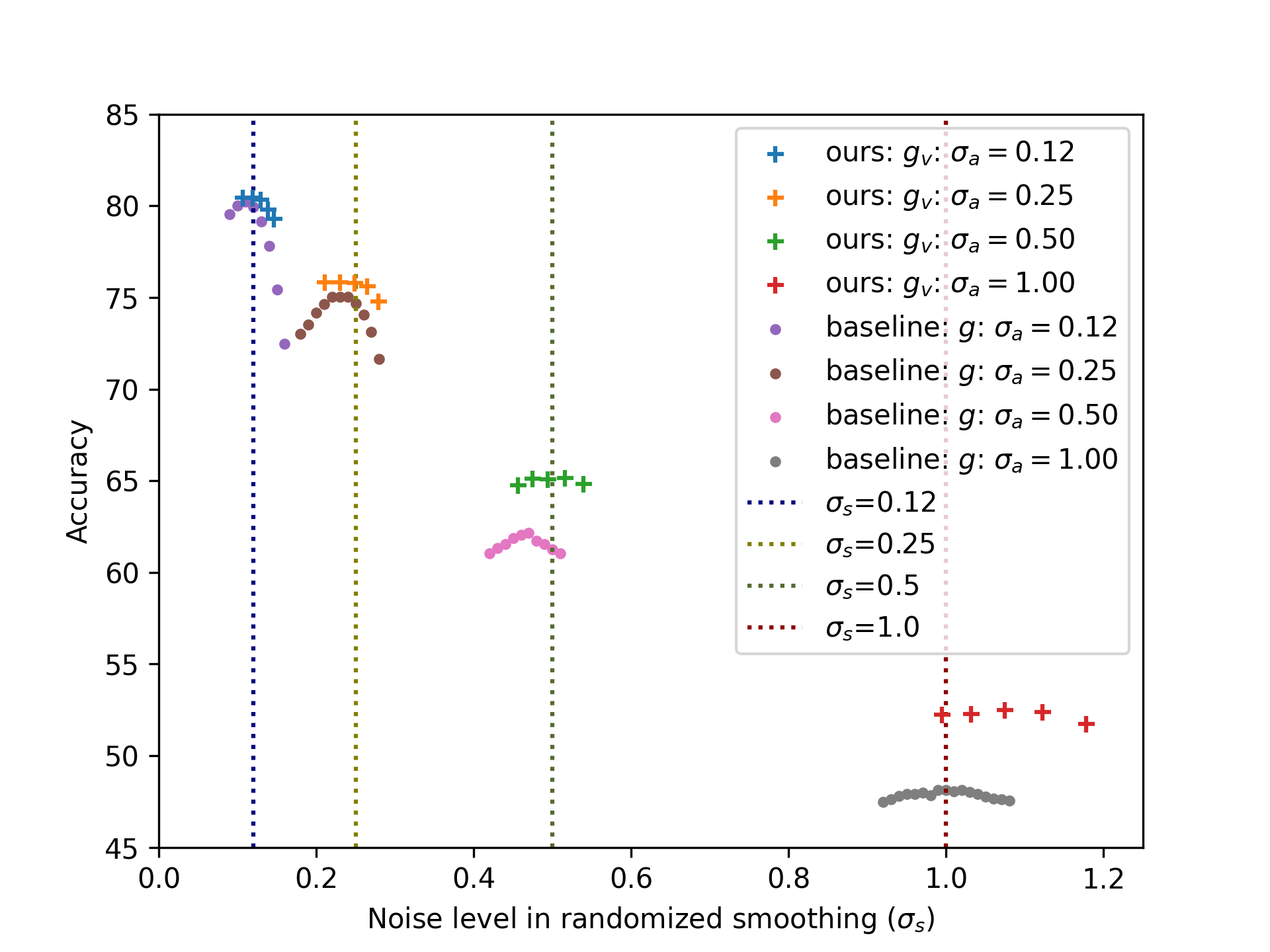}
    \caption{Clean accuracy of $g_v$ and $g$.}
    \label{fig_base_classifier_clean_input}  
  \end{subfigure}
  \hfill
  \begin{subfigure}{0.48\linewidth}
    \includegraphics[width=\columnwidth]{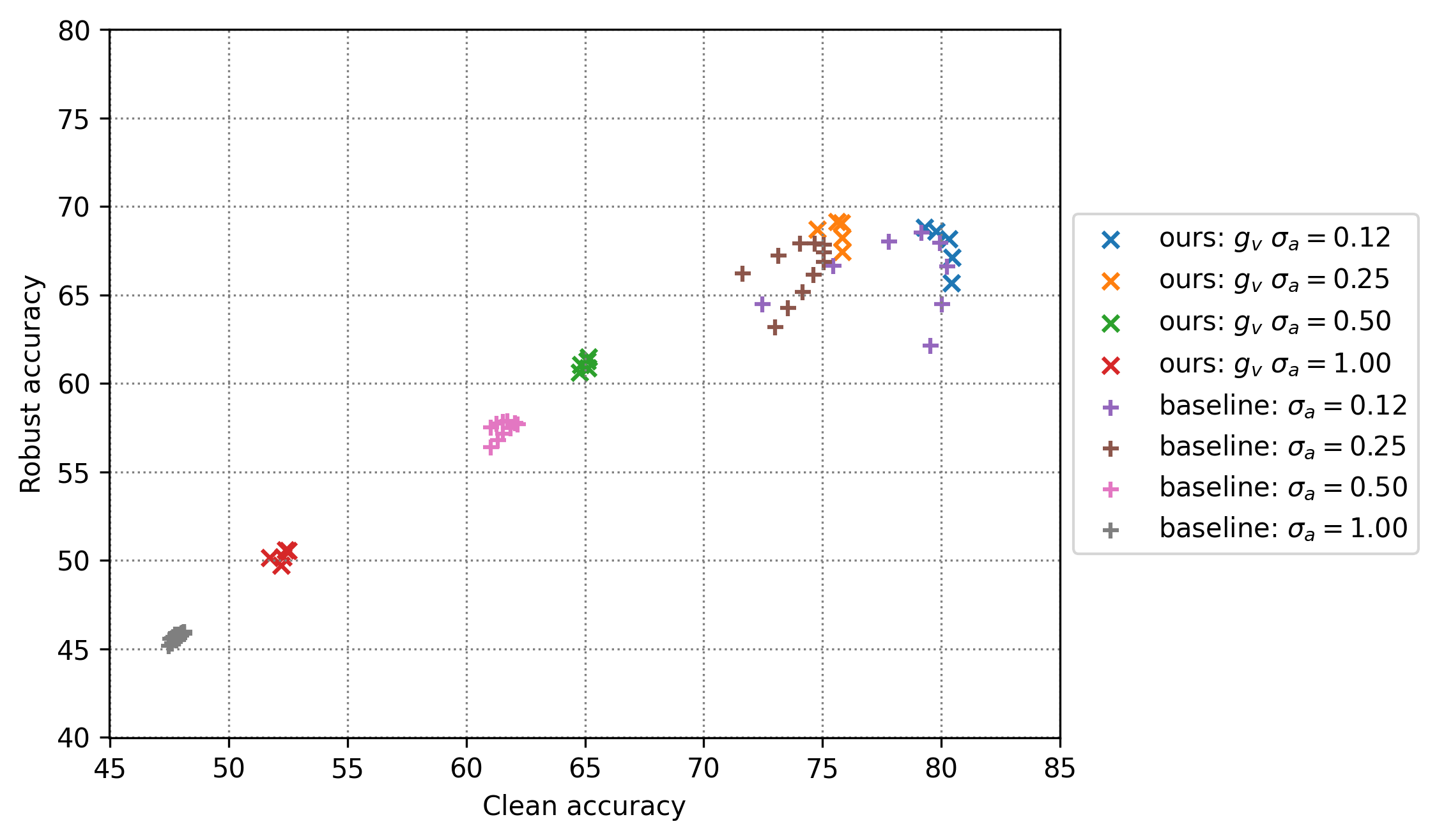}
    \caption{Clean and robust accuracy with attacked $f$.}
    \label{fig_attack_base_classifier}
  \end{subfigure}
    \caption{Clean and robust accuracy of proposed smoothing $g_v$ and baseline $g$.}
\label{fig_results_base_classifier}
\end{figure}

Fig.~\ref{fig_base_classifier_clean_input} shows the relationship between clean accuracy and noise levels $\sigma_s$ used in randomized smoothing with $g_v$ and $g$. 
For $g_v$, the mean of all selected $\sigma_s$ is plotted. 
The result suggests that selector $h$ discovered noise levels suitable for each input and contributed to better clean accuracy than the baseline.

As an example of attack results, Fig.~\ref{fig_attack_base_classifier} shows differences in clean and robust accuracy. 
The PGD attack used 200 iterations. 
Our method $g_v$ demonstrated better robustness than baseline if appropriate $\lambda$ was chosen. 
The performance of our approach is especially better for relatively larger $\sigma_a$. 
The performance of $h$ was not affected since the PGD attack did not attack $h$ in this setting.

\subsubsection{Stronger attack result:}

\begin{figure}[t]
  \centering
  \begin{subfigure}{0.48\linewidth}    
    \includegraphics[width=\columnwidth]{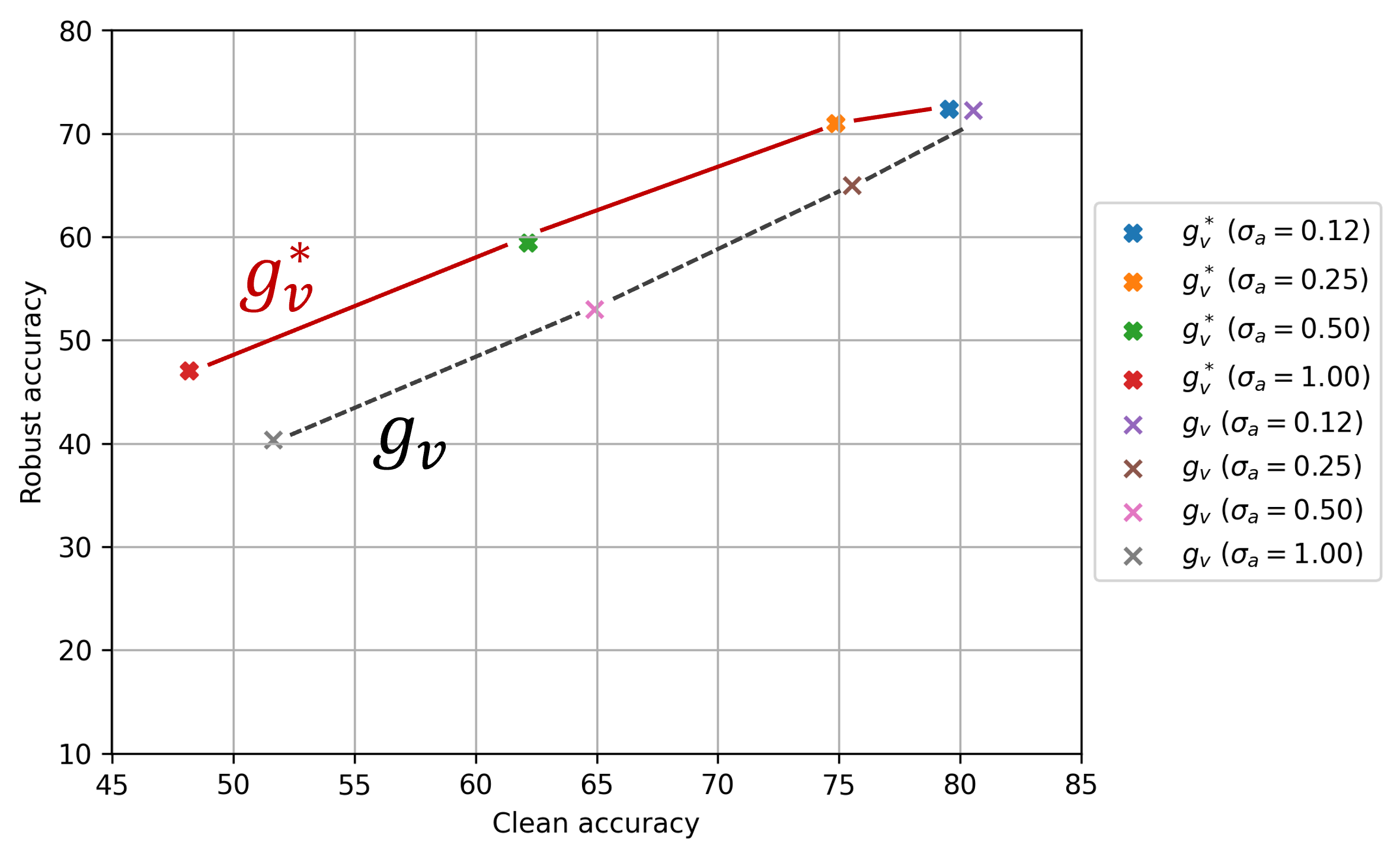}
    \caption{PGD attack with $\gamma=0.1$.}
    \label{fig_attack_smoothed_classifier_median_smoothing_effect_0.1}  
  \end{subfigure}
  \hfill
  \begin{subfigure}{0.48\linewidth}
\includegraphics[width=\columnwidth]{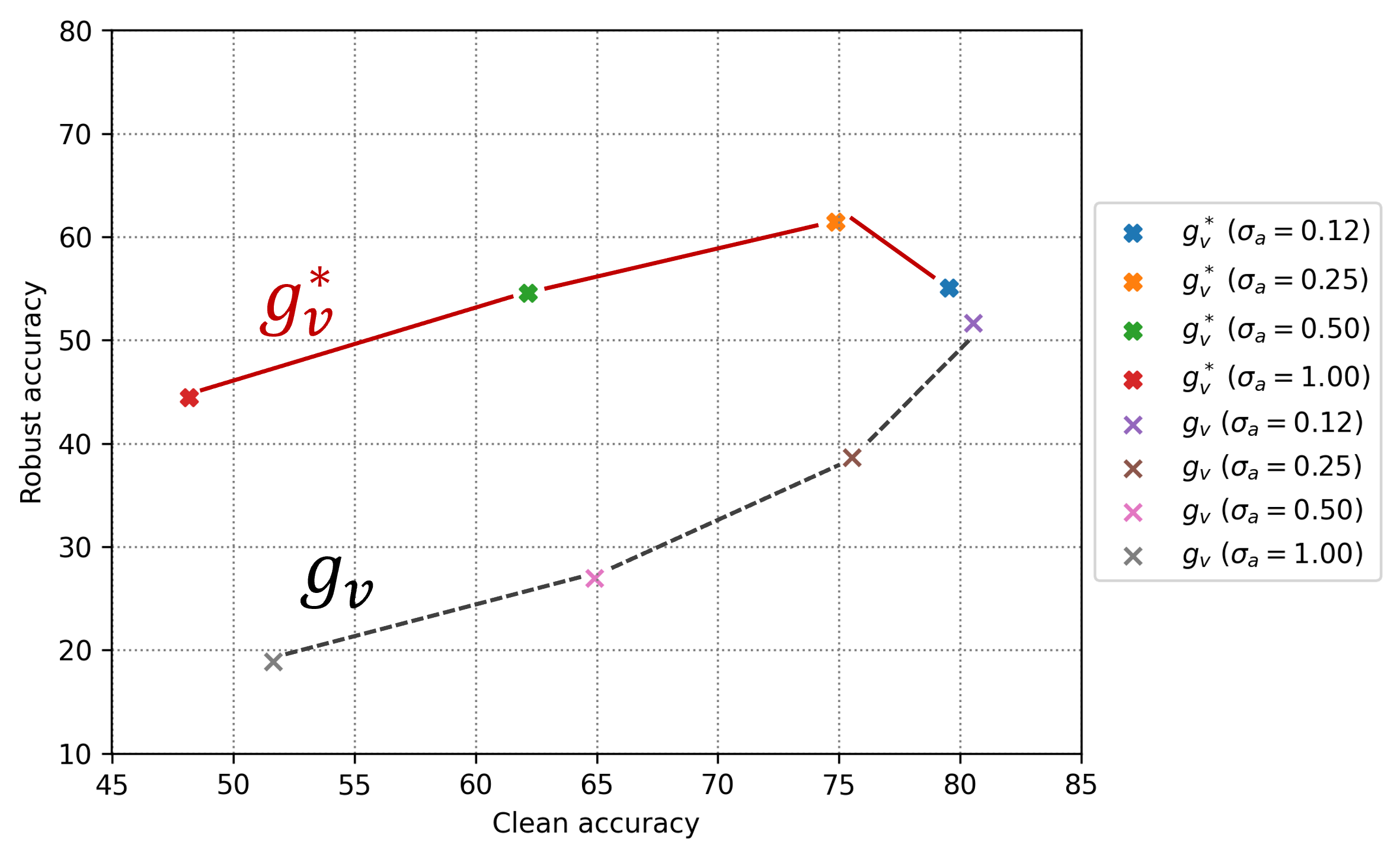}
    \caption{PGD attack with $\gamma=0.3$.}
    \label{fig_attack_smoothed_classifier_median_smoothing_effect_0.3}
  \end{subfigure}
  \caption{Attacks against variational smoothing $g_v$ and dual smoothing $g_v^*$.}
\label{fig_attack_smoothed_classifier_median_smoothing_effect}
\end{figure}

Fig.~\ref{fig_attack_smoothed_classifier_median_smoothing_effect} shows the results of clean and robust accuracy with attacks against dual-smoothing $g_v^*$ and mono-smoothing $g_v$ with $\gamma=0.1$ and $0.3$. 
The selector $h$ does not have any defensive method in $g_v$, and hence the PGD attack was effective against $h$ in this setting. 
The robust accuracy of $g_v$ dropped significantly even though $f_v$ performed well under weaker attacks as shown in Fig.~\ref{fig_attack_base_classifier}. 
On the other hand, the performance drop of $g_v^*$ was less than $g_v$, especially for larger $\gamma$. 
Although achieving better robust accuracy, one negative impact of median smoothing in $g_v^*$ is a slight degradation in the clean accuracy compared to $g_v$.

\begin{figure}[t]
  \centering
  \begin{subfigure}{0.48\linewidth}
\includegraphics[width=\columnwidth]{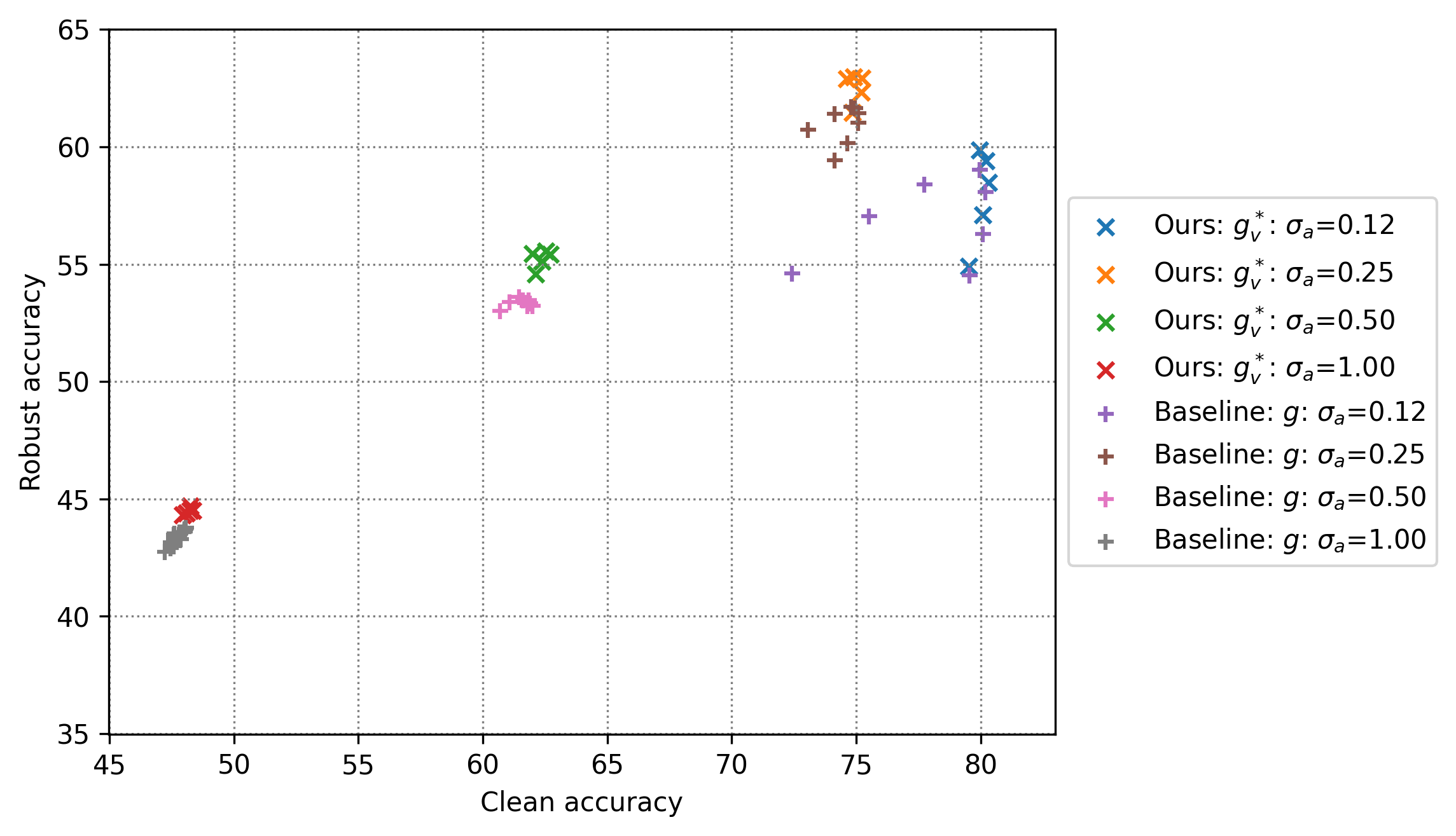}
        \caption{$g_v^*$ without clipping.}
        \label{fig_result_experimental_robustness_no_clipping}
  \end{subfigure}
  \hfill
  \begin{subfigure}{0.48\linewidth}
\includegraphics[width=\columnwidth]{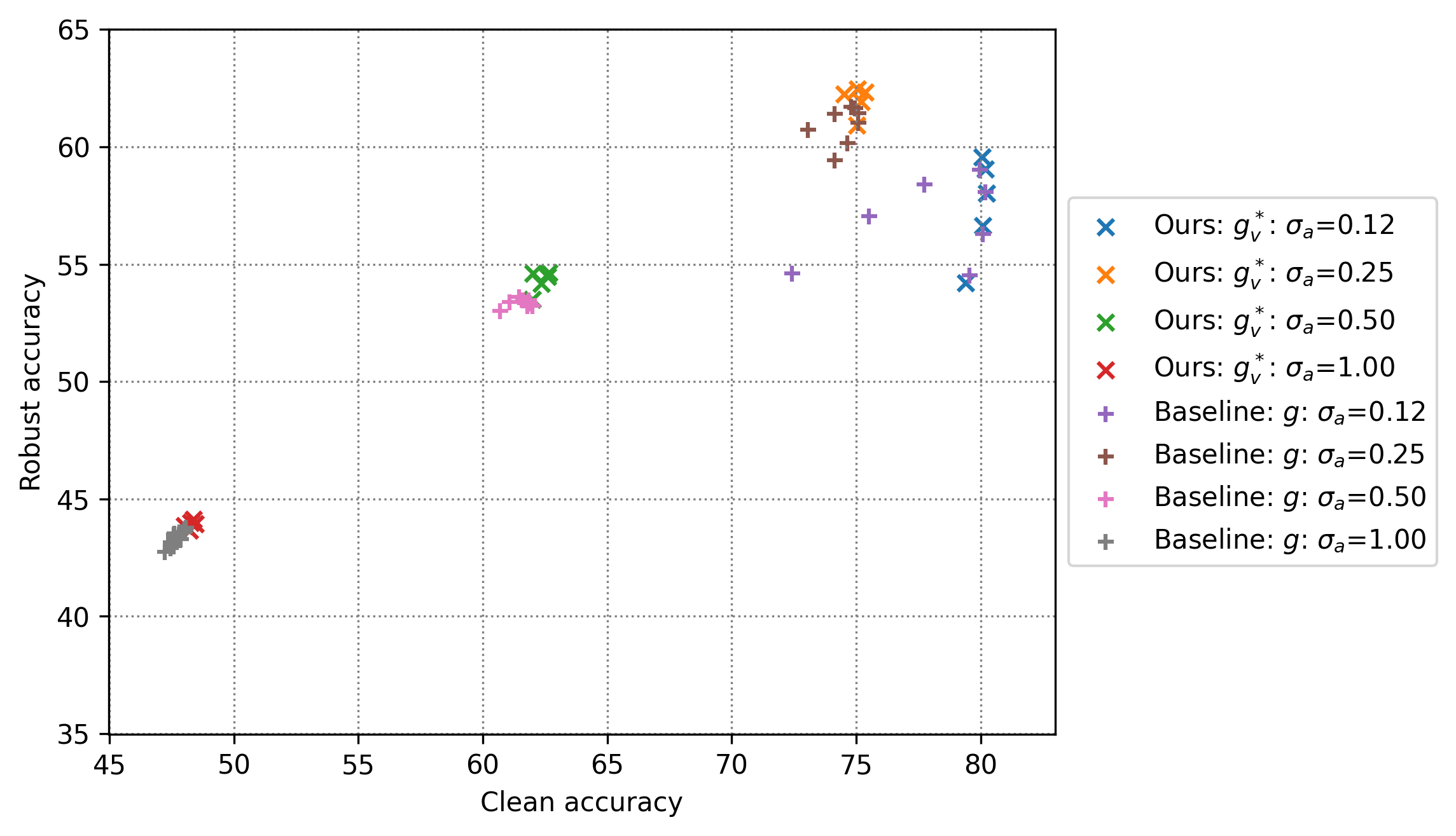}
        \caption{$g_v^*$ with clipping.}
        \label{fig_result_experimental_robustness_clipping}
  \end{subfigure}
  \caption{Clean and robust accuracy of smoothed classifier $g_v^*$.}
\label{fig_result_exp-a}
\end{figure}

Fig.~\ref{fig_result_exp-a} shows the clean and robust accuracies of $g_v^*$ and $g$. 
Fig.~\ref{fig_result_experimental_robustness_no_clipping} and~\ref{fig_result_experimental_robustness_clipping} demonstrate that clipping was effective in Experiment A. 
The PGD attack iterations was set to 500 to employ stronger attacks. 
Regardless of clipping, $g_v^*$ still demonstrated better clean and robust accuracy than the baselines. 

\begin{figure}[t]
    \centering
    \includegraphics[width=0.48\columnwidth]{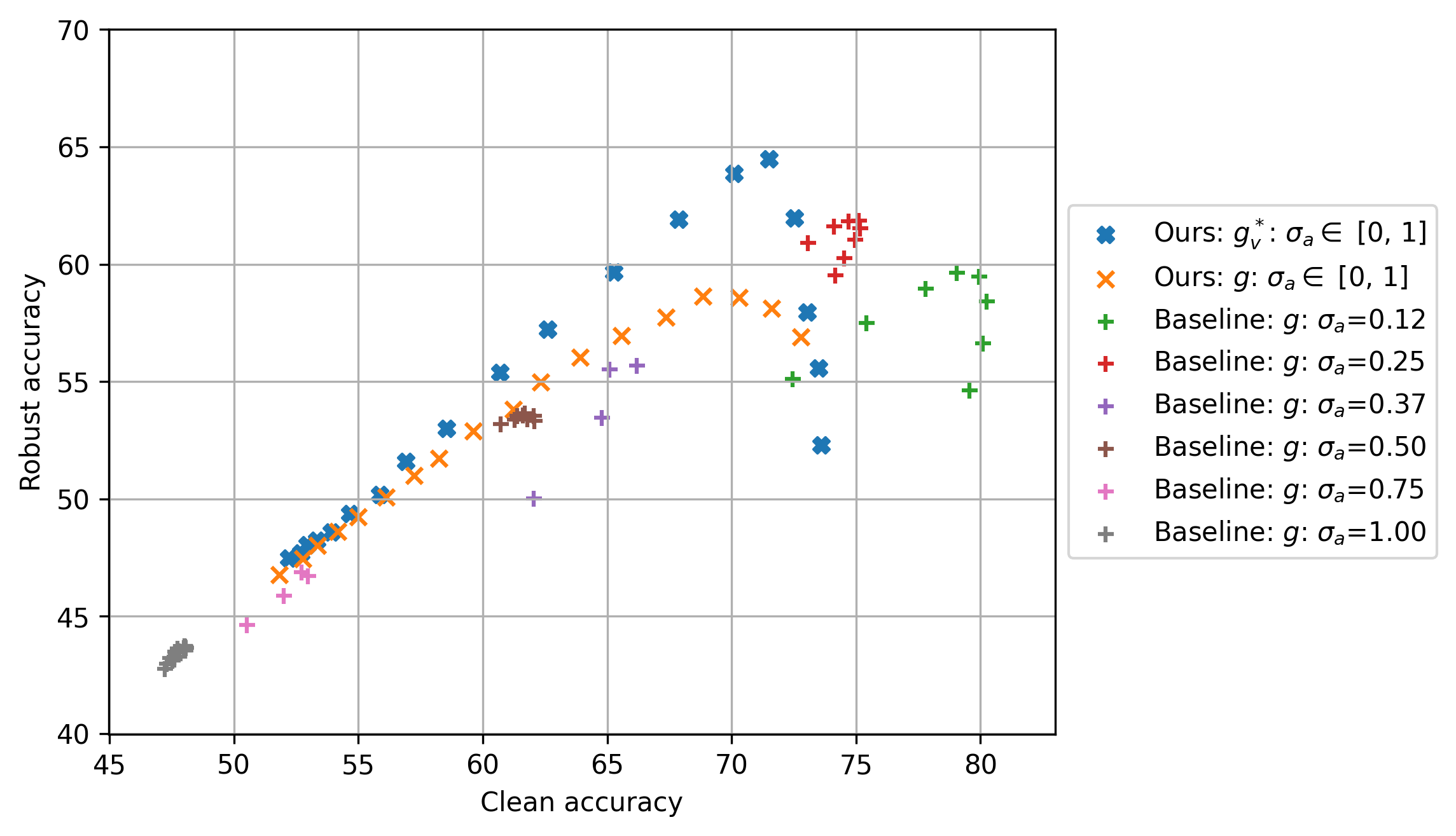}
    \caption{Attacks against smoothed classifier: universal $\sigma_a$ training.}
    \label{fig_attack_universal_sigma_a_training}
\end{figure}

For universal $\sigma_a$ training, Fig.~\ref{fig_attack_universal_sigma_a_training} shows the result of $g_v^*$ and $g$ applied to $f_u$. 
Both $g_v^*$ and $g$ demonstrated better performance with $f_u$ than $f$, especially at relatively larger $\sigma_a$. 
This result indicates $f_u$ could replace multiple $f$ in some cases. 
For instance, the $\sigma_a$ specific models $f$, for $\sigma_a\geq0.37$, could be replaced with the universal $\sigma_a$ model $f_u$ in the setting of this experiment, as shown in Fig~\ref{fig_attack_universal_sigma_a_training}.

\subsection{Discussion}
Although certified accuracy and radius of $g_v^*$ deteriorated by an attack to selector $h$, the performance of $g_v^*$ was close to the baselines when perturbation budget is relatively small. 
Furthermore, performance degradation was alleviated by applying clipping of the $h$ output. 
In terms of empirical clean and robust accuracies, our method demonstrated better performance than the baselines. 
We also showed that median smoothing and clipping applied to $g_v^*$ was effective as a defensive method for the selector $h$. 
These observations indicate practical advantages of sample-wise noise level selection using our method.

The experimental results also suggest some limitations of our method. 
Certified robustness becomes worse if the input of $h$ is attacked. 
It is harder to select better $\sigma_s$ for each input when under attack with a relatively larger distortion budget. 
In addition, as our method conducts dual randomized smoothing for $h$ and $f$, it potentially requires more computing resources since additional sampling is required for each stage of smoothing.

\section{Conclusion}
This paper proposed variational randomized smoothing, which is a framework to select noise levels suitable for each input image by using a noise level selector. 
Experimental results demonstrated enhancement of empirical robustness against adversarial attacks. 
The results also indicated certified robustness of our method is close to the levels of the baselines, when the adversarial perturbation is relatively small. 
We also showed the benefit of conditional meta learning, universal $\lambda$ training, and universal $\sigma_a$ training, so that the hyperparameters can be adjusted at test time without retraining.
These results demonstrated the advantages of sample-wise noise level selection for randomized smoothing while the employing median smoothing defense with clipping.

\bibliographystyle{splncs04}
\bibliography{main}

\newpage
\appendix
\onecolumn
\section{Theoretical and empirical bounds for percentiles}\label{appendix_theo_empi_percentiles}
\begin{figure}[h]
  \begin{subfigure}{\linewidth}    
  \centering
    \includegraphics[width=0.6\textwidth]{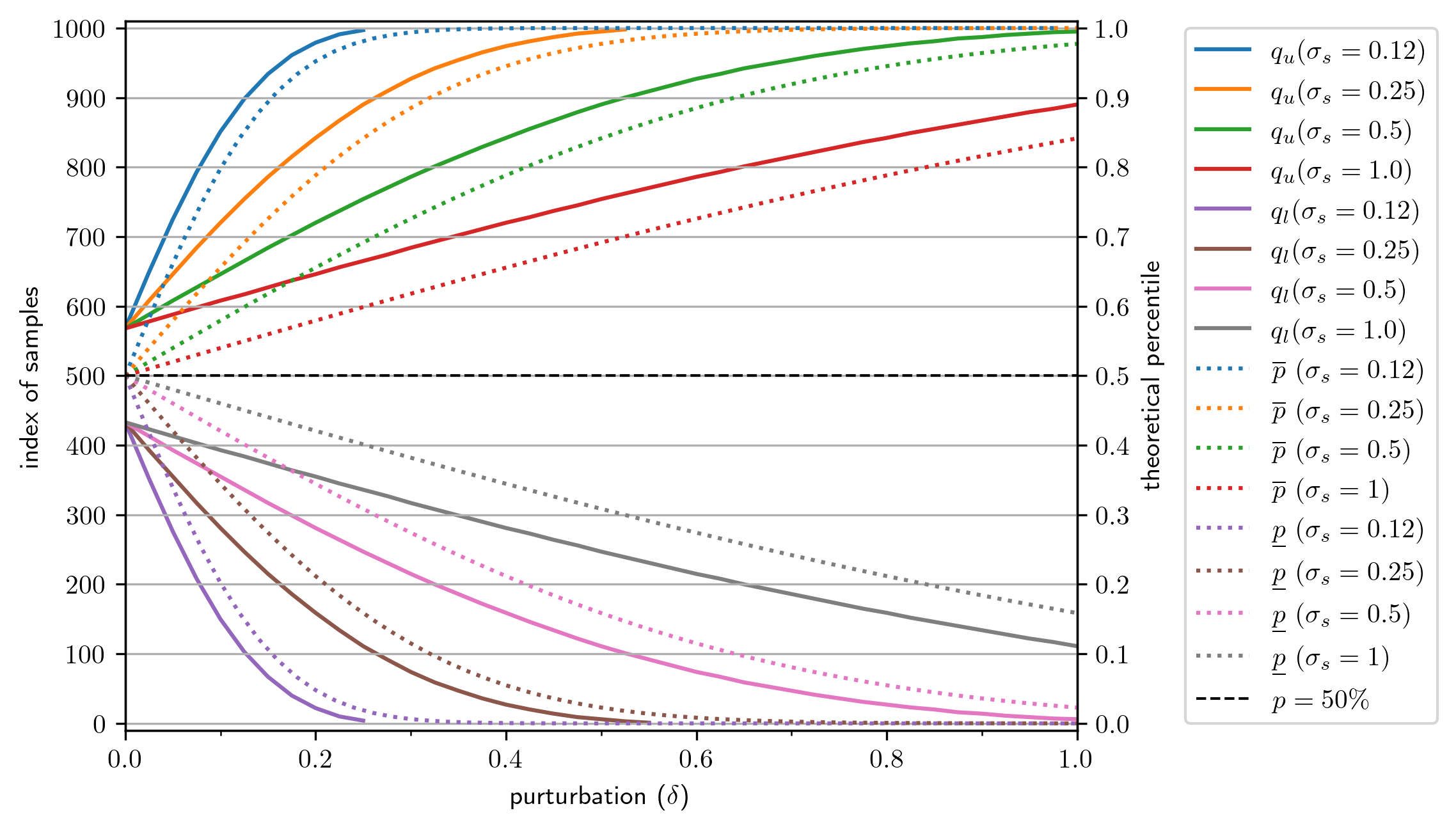}
        \caption{Percentiles ($N_h=1{,}000$)}
        \label{fig_appendix_percentiles_1000}
  \end{subfigure}\\
  \begin{subfigure}{\linewidth}
  \centering
\includegraphics[width=0.6\textwidth]{images/result_n10000.png}
        \caption{Percentiles ($N_h=10{,}000$)}
        \label{fig_appendix_percentiles_10000}
  \end{subfigure}\\
  \begin{subfigure}{\linewidth}    
  \centering
    \includegraphics[width=0.6\textwidth]{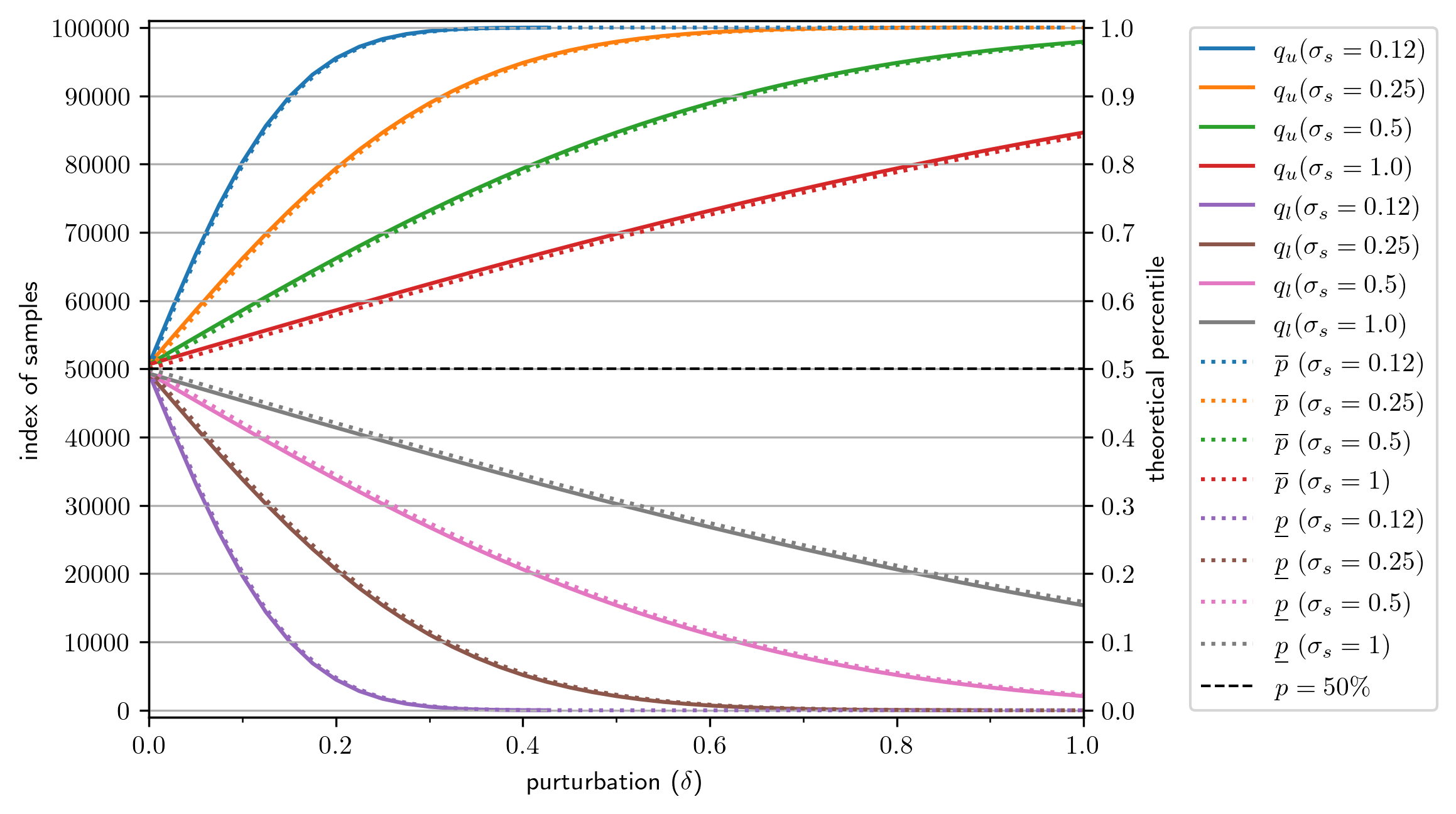}
        \caption{Percentiles ($N_h=100{,}000$)}
        \label{fig_appendix_percentiles_100000}
  \end{subfigure}
  \caption{Theoretical and empirical bounds for median smoothing percentiles.}
\label{fig_appendix_percentile_bounds}
\end{figure}

Fig.~\ref{fig_appendix_percentile_bounds} shows the theoretical and empirical upper and lower bounds for the percentiles used in median smoothing.
The theoretical upper and lower bounds of percentiles are respectively denoted by $\overline{p}$ and $\underline{p}$.
The empirical upper and lower bounds of percentiles are respectively denoted by $q_u$ and $q_l$, which correspond to the indices of the $N_h$ samples of the noise level selector $h$ output.
The theoretical and empirical bounds get closer as $N_h$ becomes larger as shown in~Fig. \ref{fig_appendix_percentile_bounds}.

\clearpage

\section{Empirical upper and lower bounds for selector output}\label{appendix_empirical_sigma_bounds}
\begin{figure}[h]
  \centering
  \begin{subfigure}{0.45\linewidth}    
    \includegraphics[width=\columnwidth]{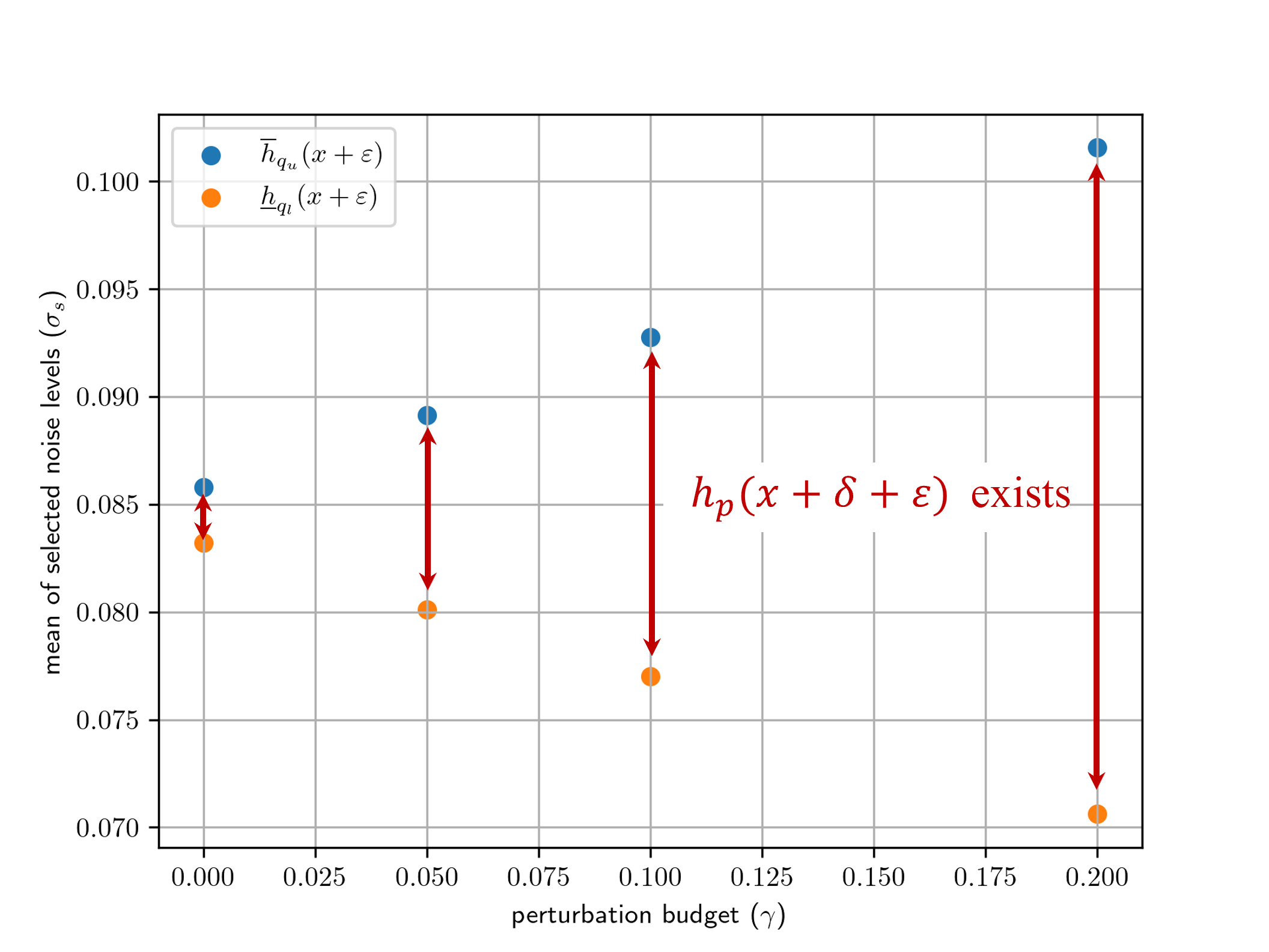}
        \caption{$h$ for $f$ trained with $\sigma_a=0.12$.}
        \label{fig_appendix_experimental_bounds_examples_0.12}
  \end{subfigure}
  \hfill
  \begin{subfigure}{0.45\linewidth}
\includegraphics[width=\columnwidth]{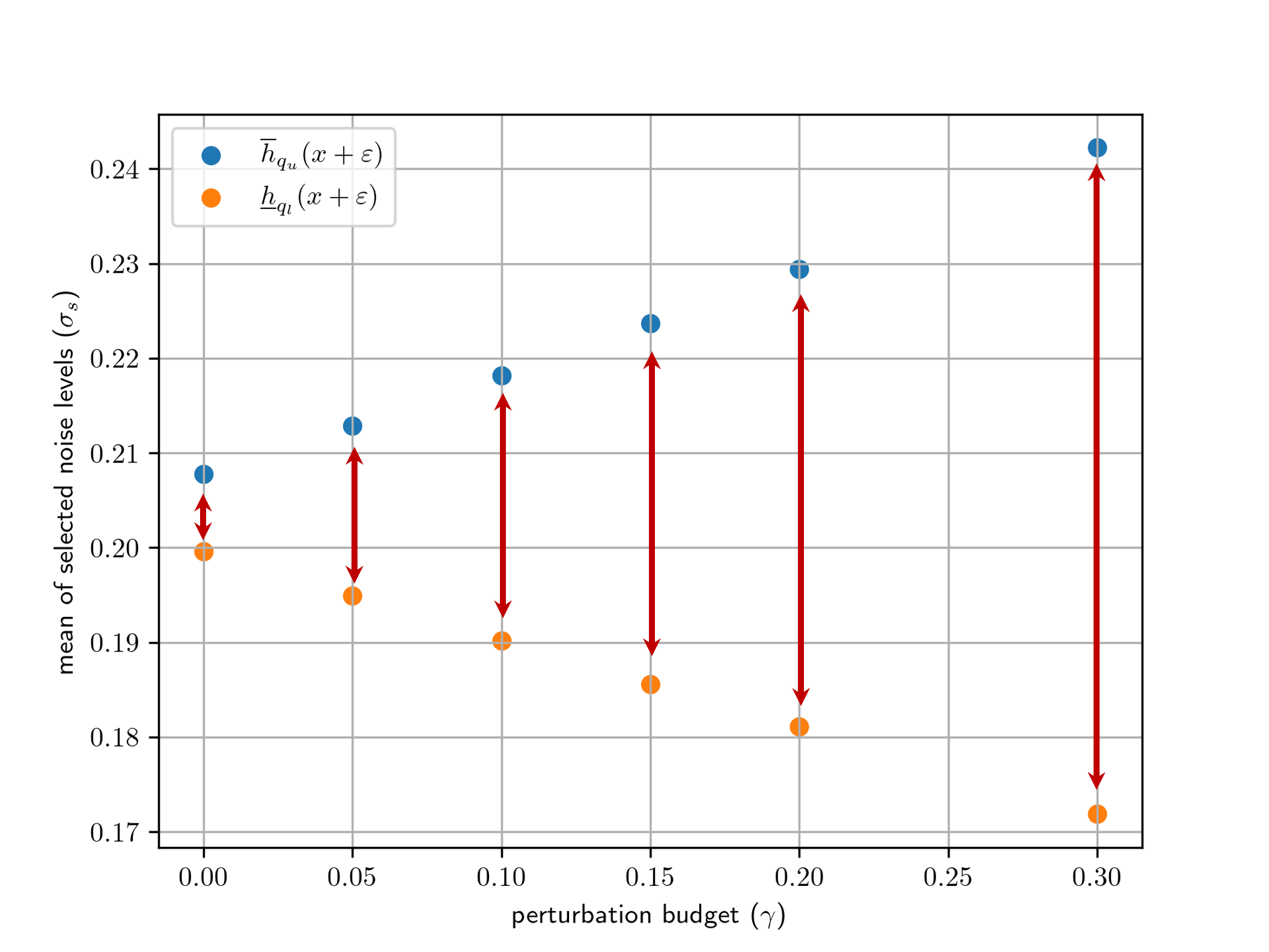}
        \caption{$h$ for $f$ trained with $\sigma_a=0.25$.}
        \label{fig_appendix_experimental_bounds_examples_0.25}
  \end{subfigure}\\
  \begin{subfigure}{0.45\linewidth}    
    \includegraphics[width=\columnwidth]{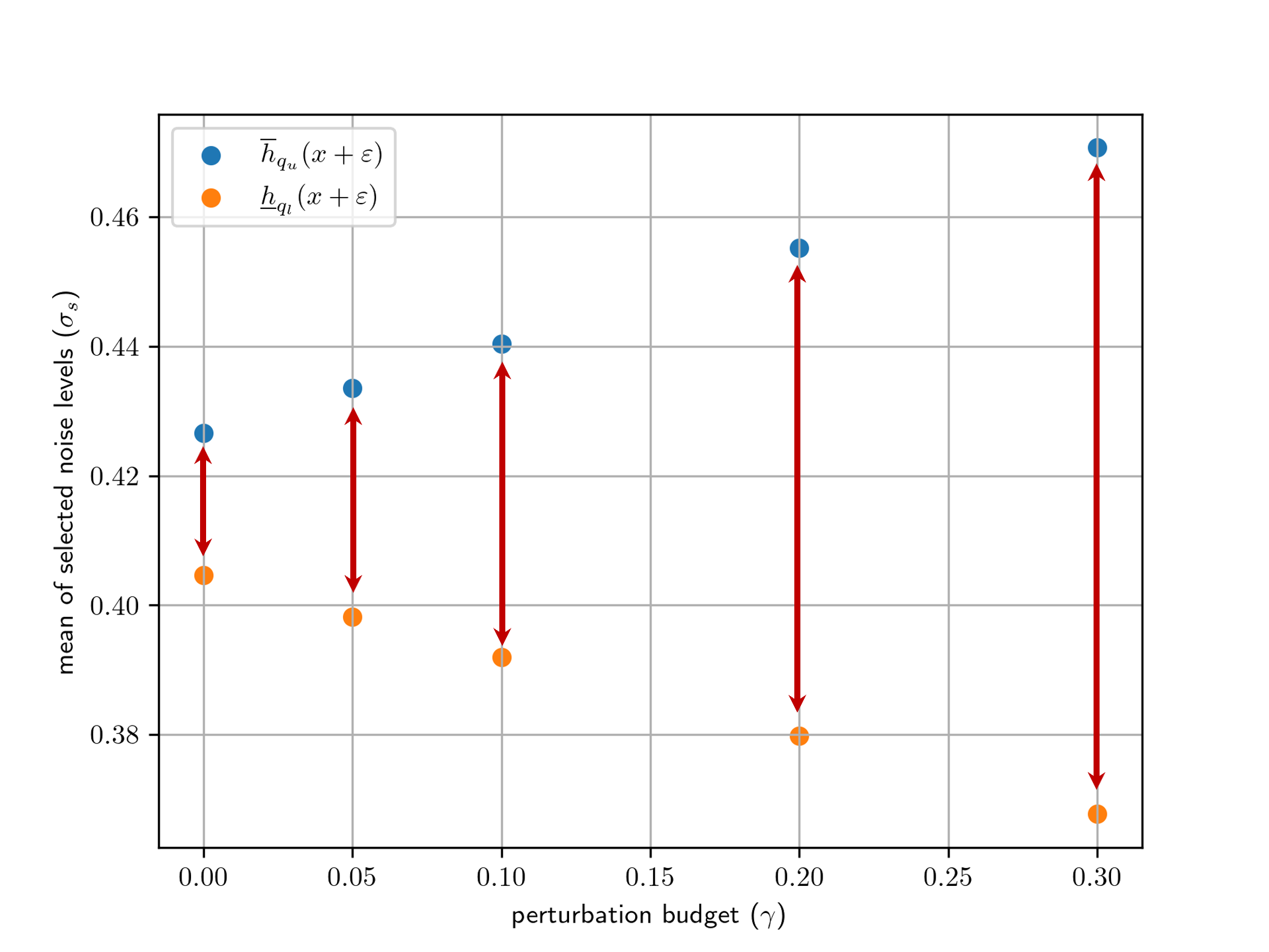}
        \caption{$h$ for $f$ trained with $\sigma_a=0.50$.}
        \label{fig_appendix_experimental_bounds_examples_0.50}
  \end{subfigure}
  \hfill
  \begin{subfigure}{0.45\linewidth}
\includegraphics[width=\columnwidth]{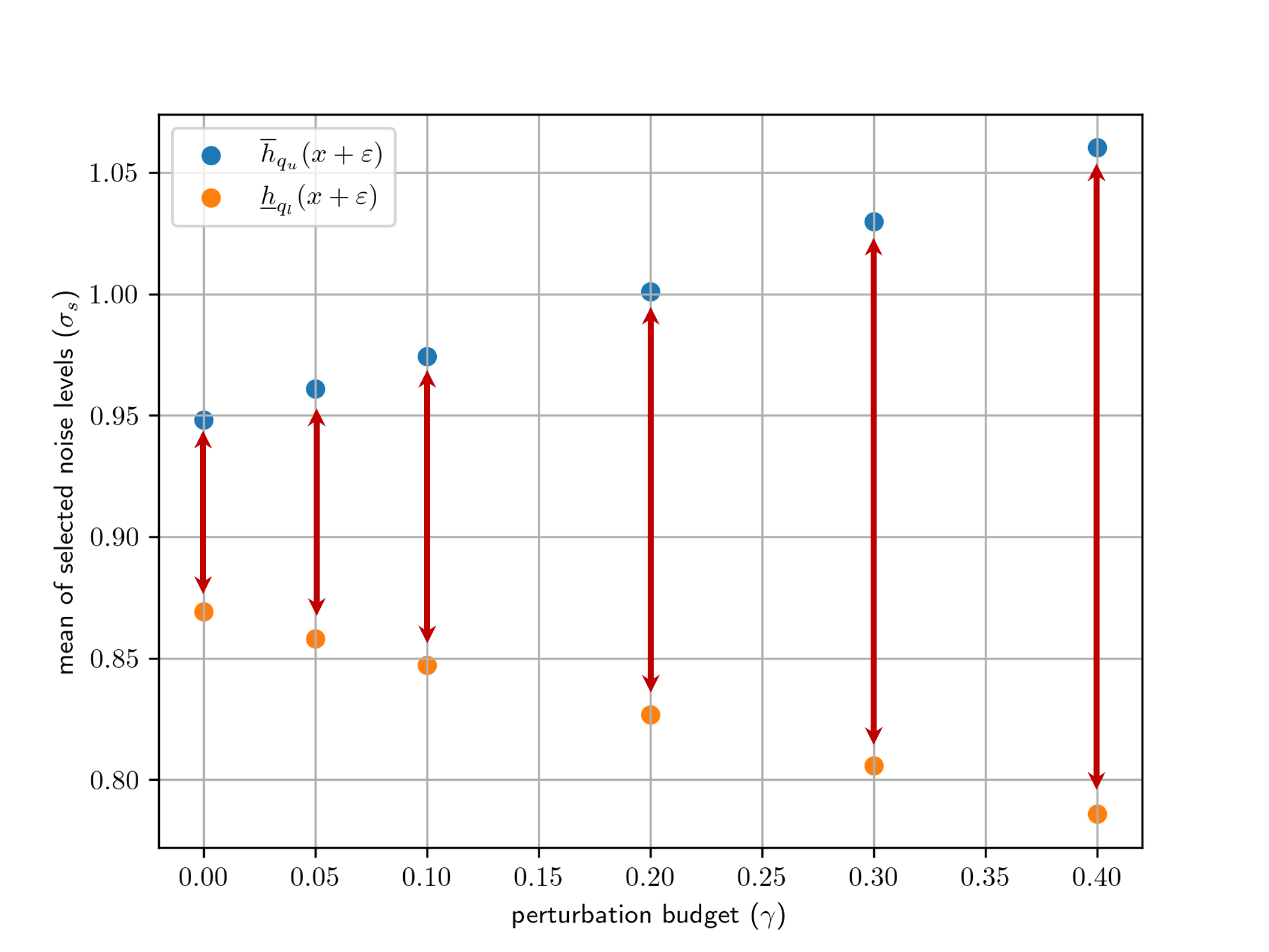}
        \caption{$h$ for $f$ trained with $\sigma_a=1.00$.}
        \label{fig_appendix_experimental_bounds_examples_1.00}
  \end{subfigure}
  \caption{Empirical upper and lower bounds for selector output.}
\label{fig_appendix_experimental_bounds}
\end{figure}

Fig.~\ref{fig_appendix_experimental_bounds} shows the empirical upper and lower bounds for the selector output $h_p(x+\delta+\varepsilon)$. 
The upper and lower bounds are denoted by $\overline{h}_{q_u}(x+\varepsilon)$ and $\underline{h}_{q_l}(x+\varepsilon)$, respectively. 
The test set of CIFAR-10 dataset was used as the input for $h$. 
The parameters $N_h=1{,}000$ and $\lambda=0.0$ were used for $h$ across all settings. 
$\overline{h}_{q_u}(x+\varepsilon)$ and $\underline{h}_{q_l}(x+\varepsilon)$ were averaged over all images in the test set. 
This result suggests that deviations of $\overline{h}_{q_u}(x+\varepsilon)$ and $\underline{h}_{q_l}(x+\varepsilon)$ from $\underline{h}_p(x+\varepsilon)$, which is the optimal value for median smoothing, could be larger as adversarial distortion budget $\gamma$ increases.

\clearpage

\section{Extensive analysis of clean and robust accuracy with base classifier trained by universal $\sigma_a$ training}\label{appendix_universal_sigma_training}
Regarding smoothed classifier $g$, tradeoff between clean and robust accuracy can be analyzed by sweeping $\sigma_s$ against base classifier $f$. Likewise, $g_v^*$ demonstrates different clean and robust accuracy by sweeping $\lambda$. The performance of a smoothed classifier using the base classifier $f_u$ also depends on the $\sigma_a'$ chosen for universal $\sigma_a$ training. Although Fig.~\ref{fig_attack_universal_sigma_a_training} shows one example of clean and robust accuracy with the case where $\sigma_a'=1.0$ and $\gamma=0.3$, different curves about tradeoff between clean and robust accuracy can be seen with different sets of $\sigma_a'$ and $\gamma$. 
Hence, we conducted extensive experiments similar to the experimental results shown in Fig.~\ref{fig_attack_universal_sigma_a_training}. 
In these experiments, $\sigma_a'$ was chosen from one of $0.25$, $0.50$, and $1.00$. 
For training, $h$ was conditioned on $\sigma_a=\sigma_a'/2$ and 
$\sigma_t$ was set to $\sigma_a'$. 
To analyze differences across the perturbation budget of the PGD attack, we varied $\gamma$ from $0.1$, $0.3$, and $0.5$. 
For experiments of $g_v^*$, $\lambda$ was selected from $\{0.0, 0.1, \ldots, 0.5\}$ except the case where $\sigma_a'=0.25$ and $\gamma=0.5$. 
For the case with $\sigma_a'=0.25$ and $\gamma=0.5$, $\lambda$ was chosen from $\{0.0, 0.1, \ldots, 0.9\}$.

\begin{figure}[h]
  \centering
  \begin{subfigure}{0.45\linewidth}
\includegraphics[width=\columnwidth]{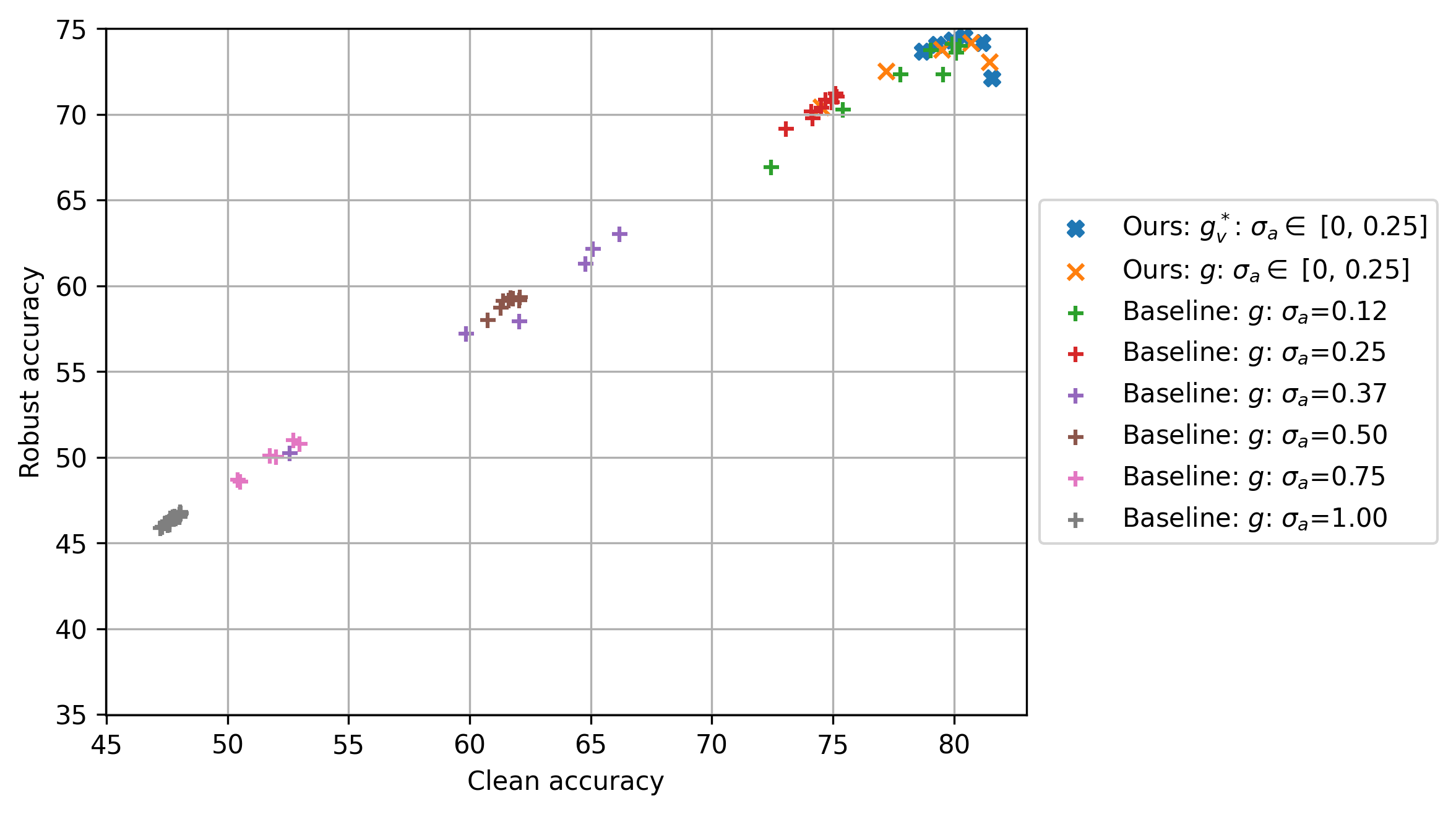}
        \caption{Perturbation budget for robust accuracy is $\gamma=0.1$.}
        \label{fig_appendix_experimental_bounds_examples_sigma0to0.25_gamma0.1}
  \end{subfigure}\\
  \begin{subfigure}{0.45\linewidth}    
    \includegraphics[width=\columnwidth]{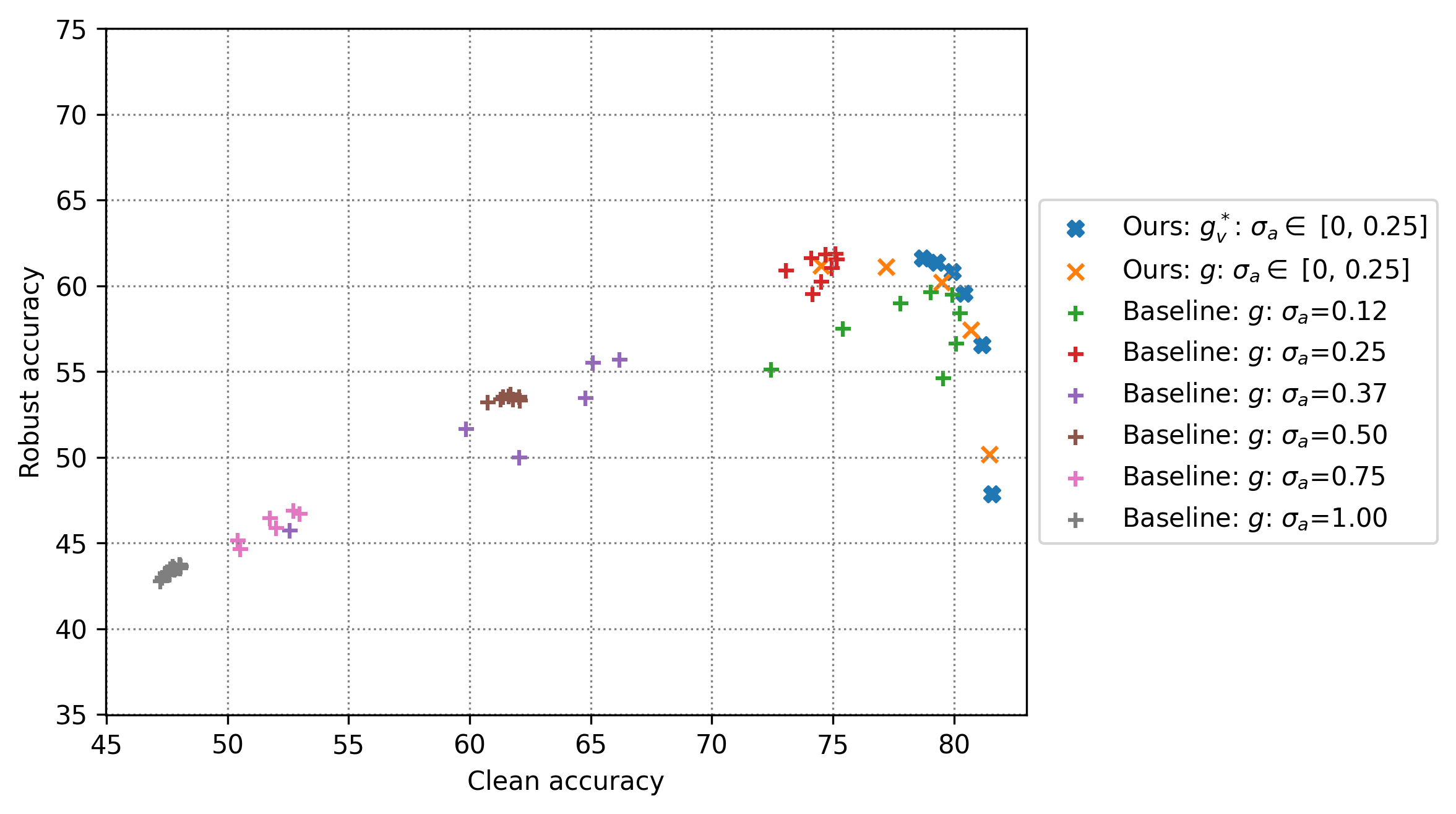}
        \caption{Perturbation budget for robust accuracy is $\gamma=0.3$.}
        \label{fig_appendix_experimental_bounds_examples_sigma0to0.25_gamma0.3}
  \end{subfigure}
  \hfill
  \begin{subfigure}{0.45\linewidth}
\includegraphics[width=\columnwidth]{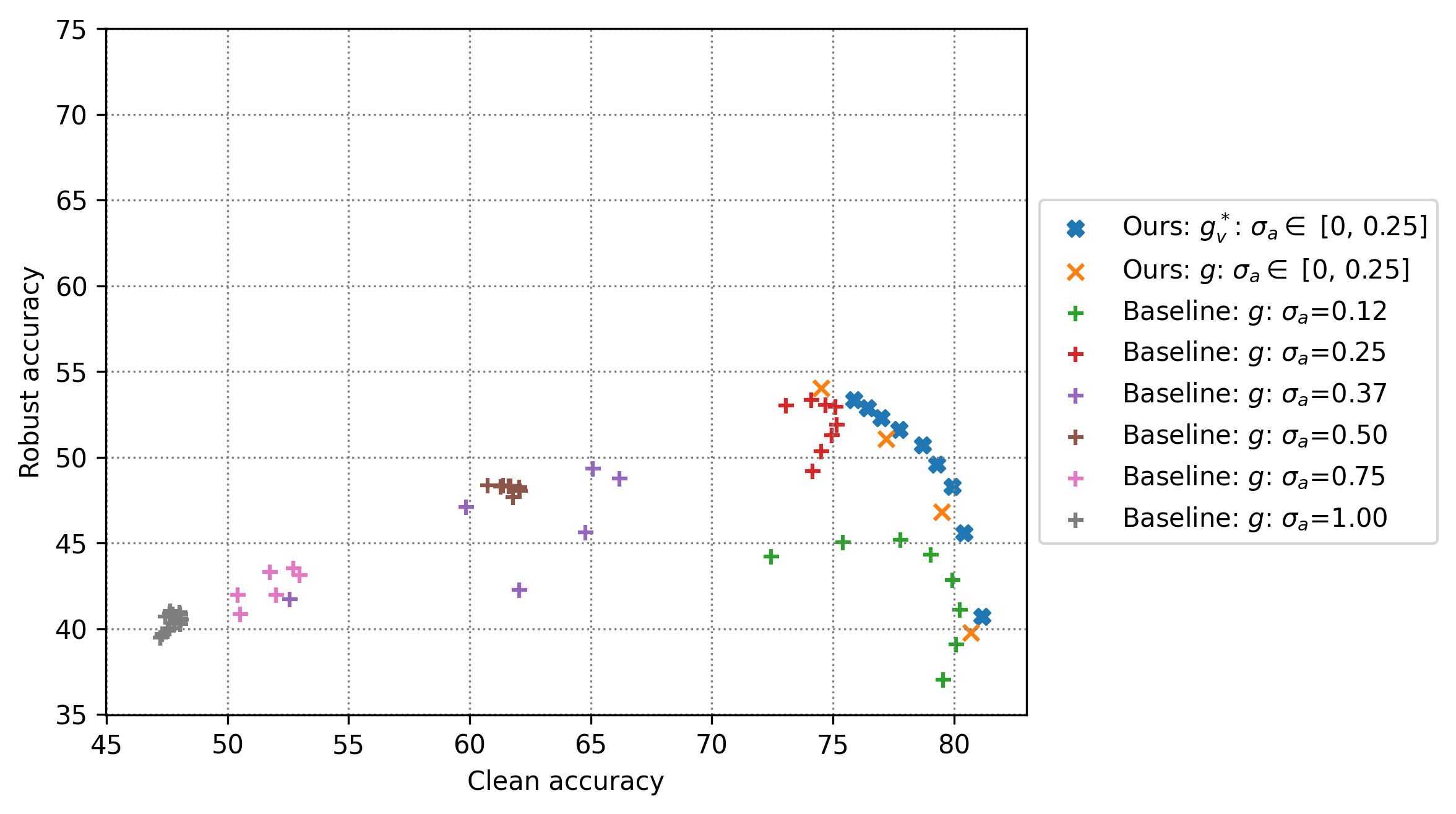}
        \caption{Perturbation budget for robust accuracy is $\gamma=0.5$.}
        \label{fig_appendix_experimental_bounds_examples_sigma0to0.25_gamma0.5}
  \end{subfigure}
  \caption{Clean and robust accuracy: fixed $\sigma_a$ training versus universal $\sigma_a$ training ($\sigma_a'=0.25$).}
\label{fig_appendix_universal_sigma_training_sigma0.25}
\end{figure}

\begin{figure}[h]
  \centering
  \begin{subfigure}{0.45\linewidth}
\includegraphics[width=\columnwidth]{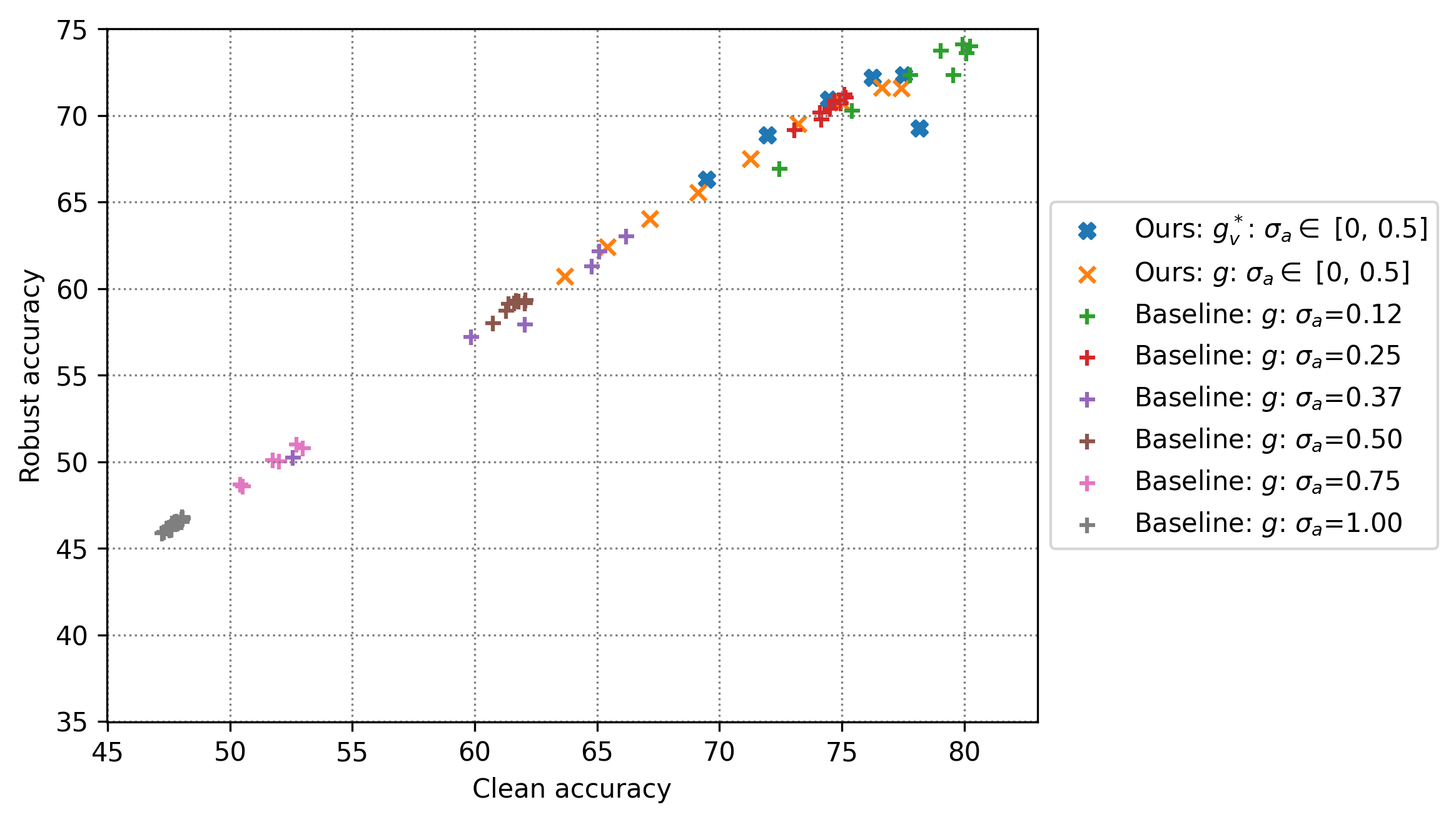}
        \caption{Perturbation budget for robust accuracy is $\gamma=0.1$.}
        \label{fig_appendix_experimental_bounds_examples_sigma0to0.50_gamma0.1}
  \end{subfigure}\\
  \begin{subfigure}{0.45\linewidth}    
    \includegraphics[width=\columnwidth]{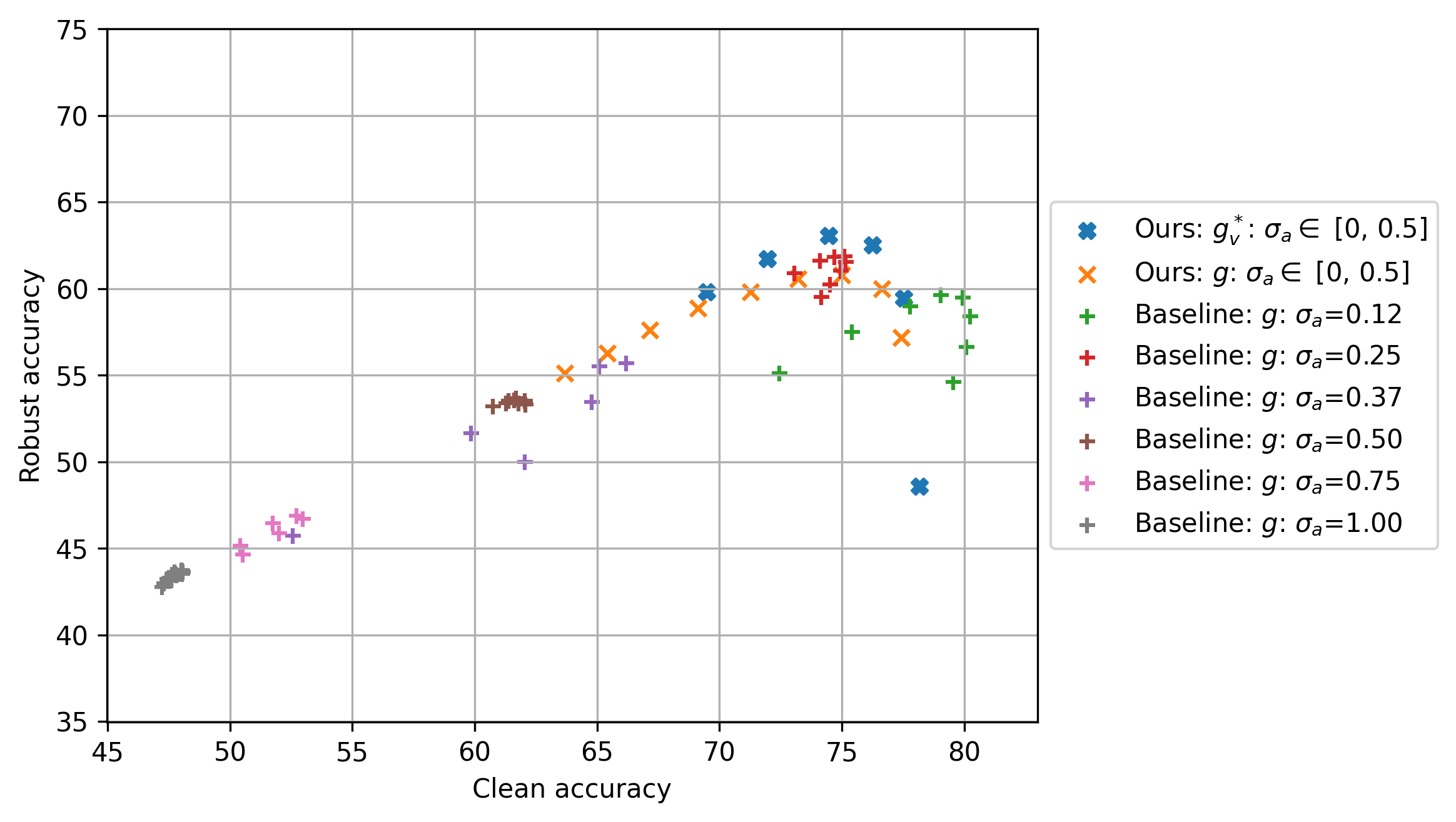}
        \caption{Perturbation budget for robust accuracy is $\gamma=0.3$.}
        \label{fig_appendix_experimental_bounds_examples_sigma0to0.50_gamma0.3}
  \end{subfigure}
  \hfill
  \begin{subfigure}{0.45\linewidth}
\includegraphics[width=\columnwidth]{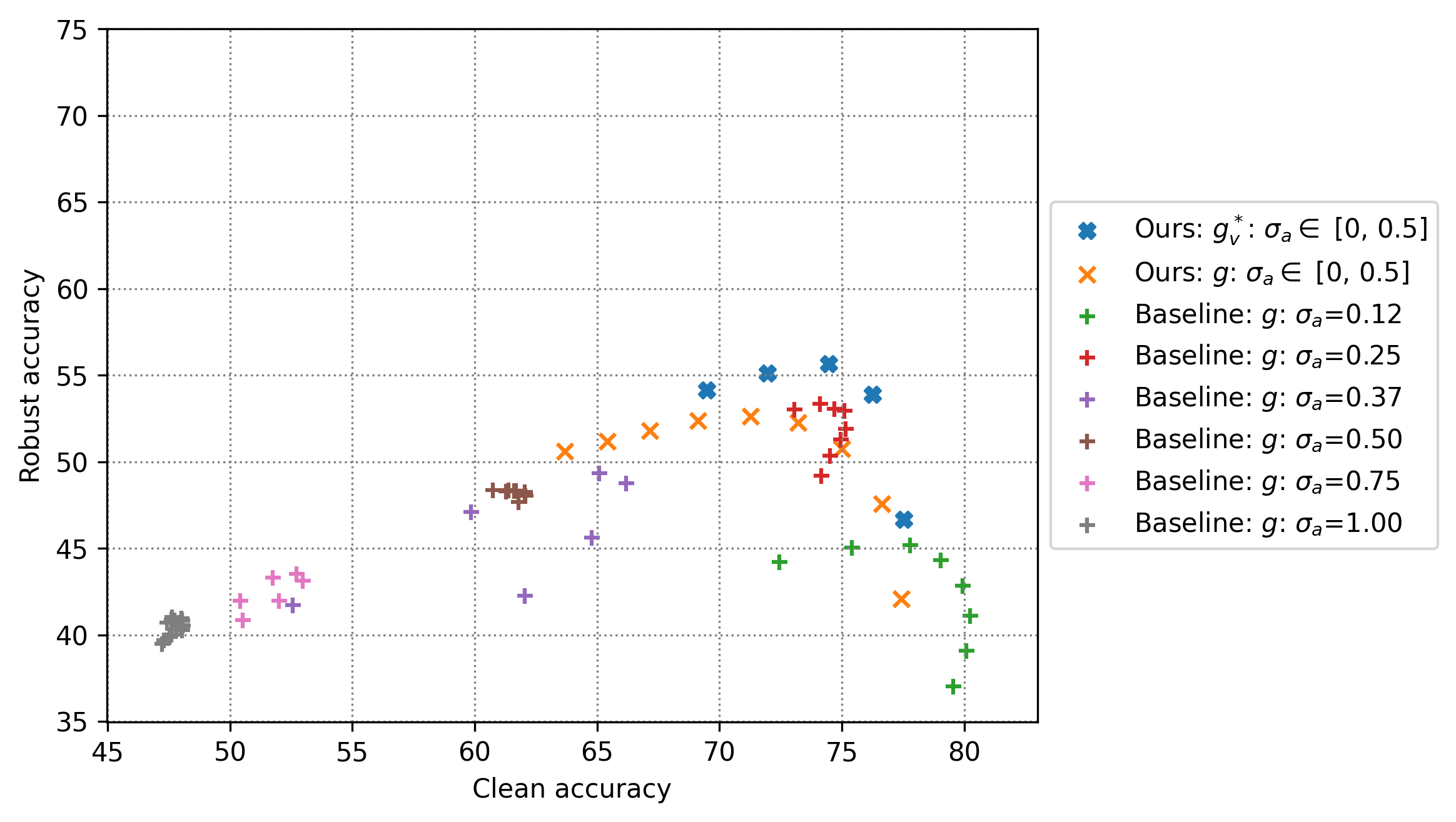}
        \caption{Perturbation budget for robust accuracy is $\gamma=0.5$.}
        \label{fig_appendix_experimental_bounds_examples_sigma0to0.50_gamma0.5}
  \end{subfigure}
  \caption{Clean and robust accuracy: fixed $\sigma_a$ training versus universal $\sigma_a$ training ($\sigma_a'=0.50$).}
\label{fig_appendix_universal_sigma_training_sigma0.50}
\end{figure}

\begin{figure}[h]
  \centering
  \begin{subfigure}{0.45\linewidth}
\includegraphics[width=\columnwidth]{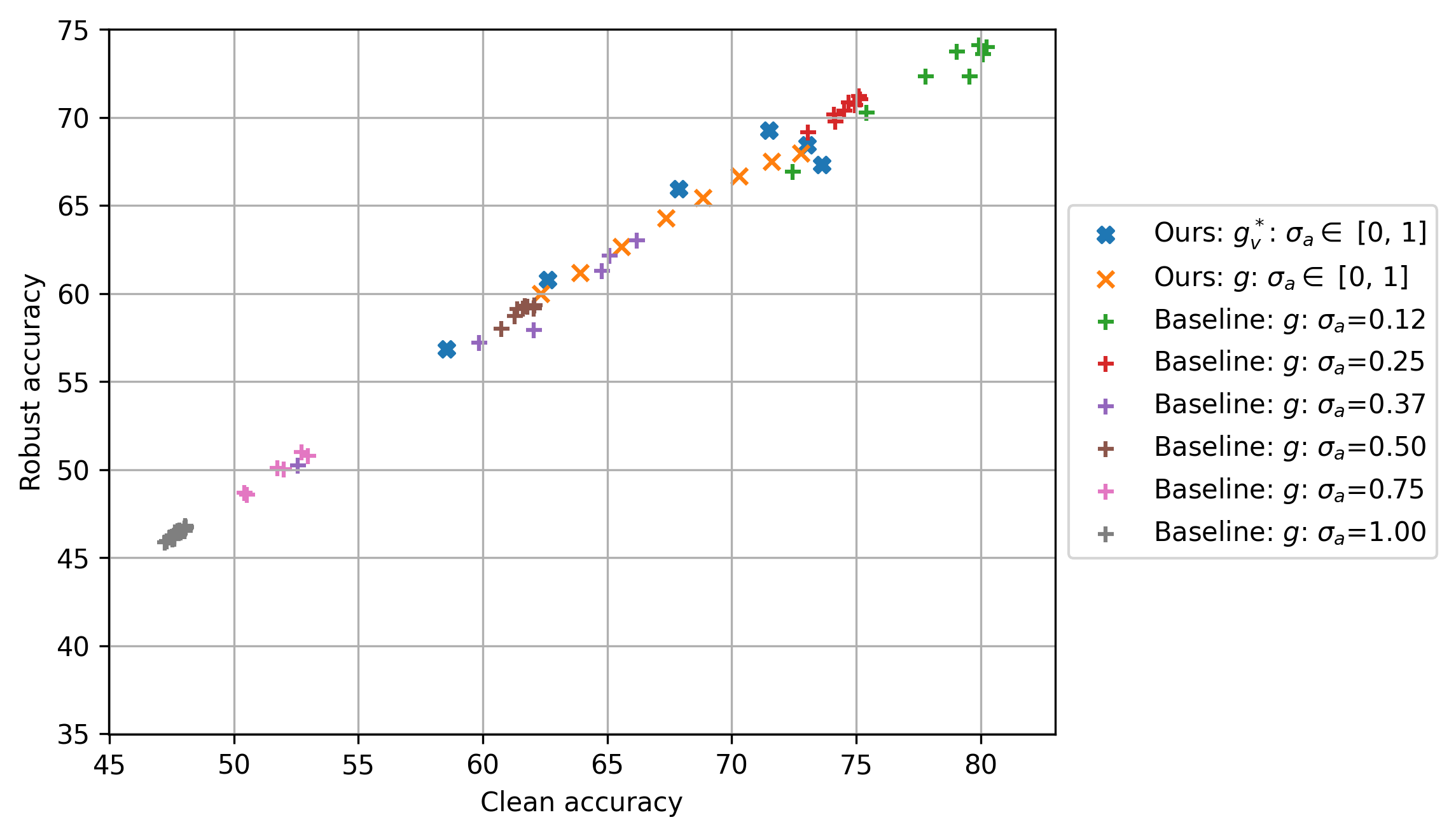}
        \caption{Perturbation budget for robust accuracy is $\gamma=0.1$.}
        \label{fig_appendix_experimental_bounds_examples_sigma0to1.00_gamma0.1}
  \end{subfigure}\\
  \begin{subfigure}{0.45\linewidth}    
    \includegraphics[width=\columnwidth]{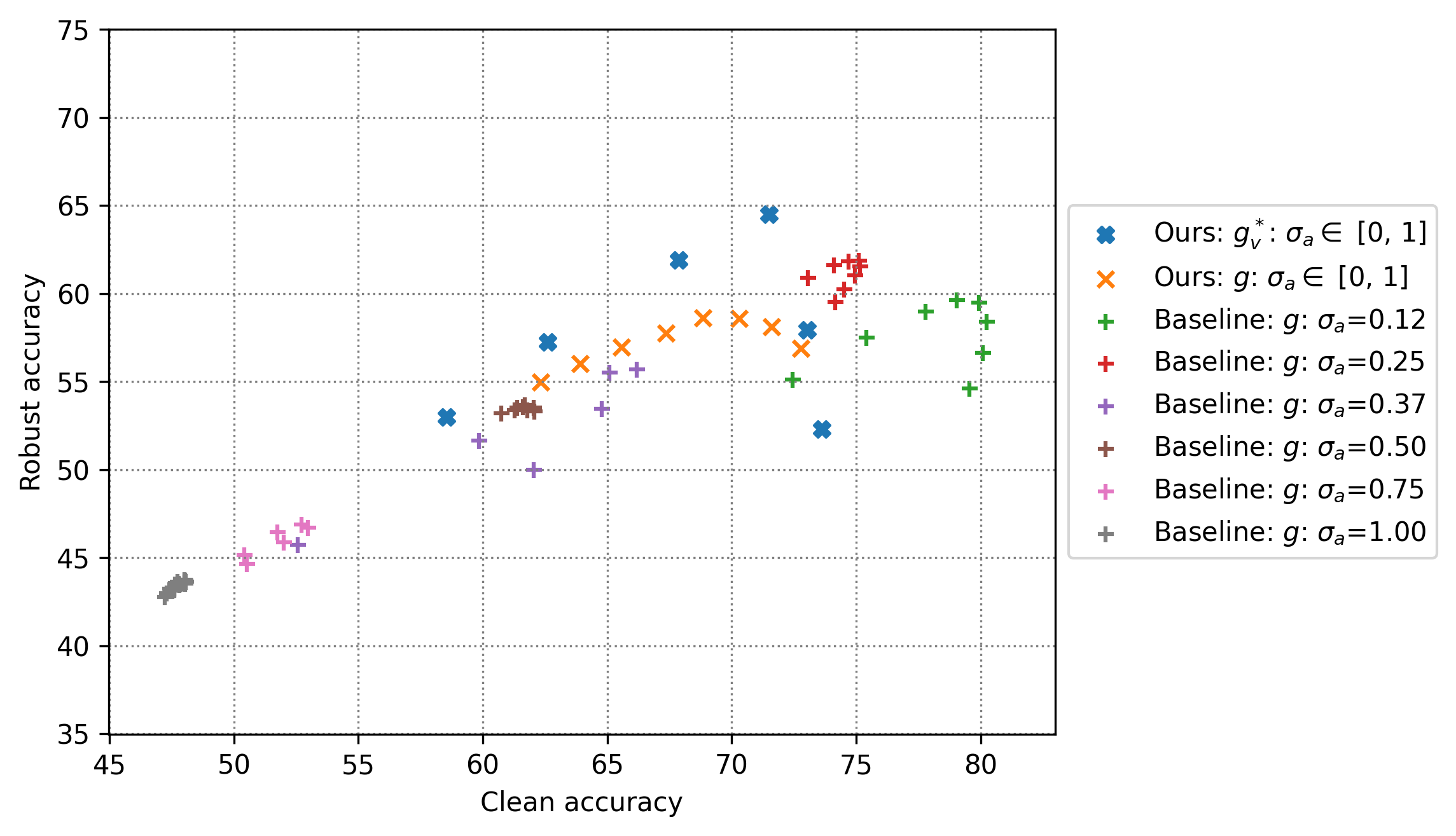}
        \caption{Perturbation budget for robust accuracy is $\gamma=0.3$.}
        \label{fig_appendix_experimental_bounds_examples_sigma0to1.00_gamma0.3}
  \end{subfigure}
  \hfill
  \begin{subfigure}{0.45\linewidth}
\includegraphics[width=\columnwidth]{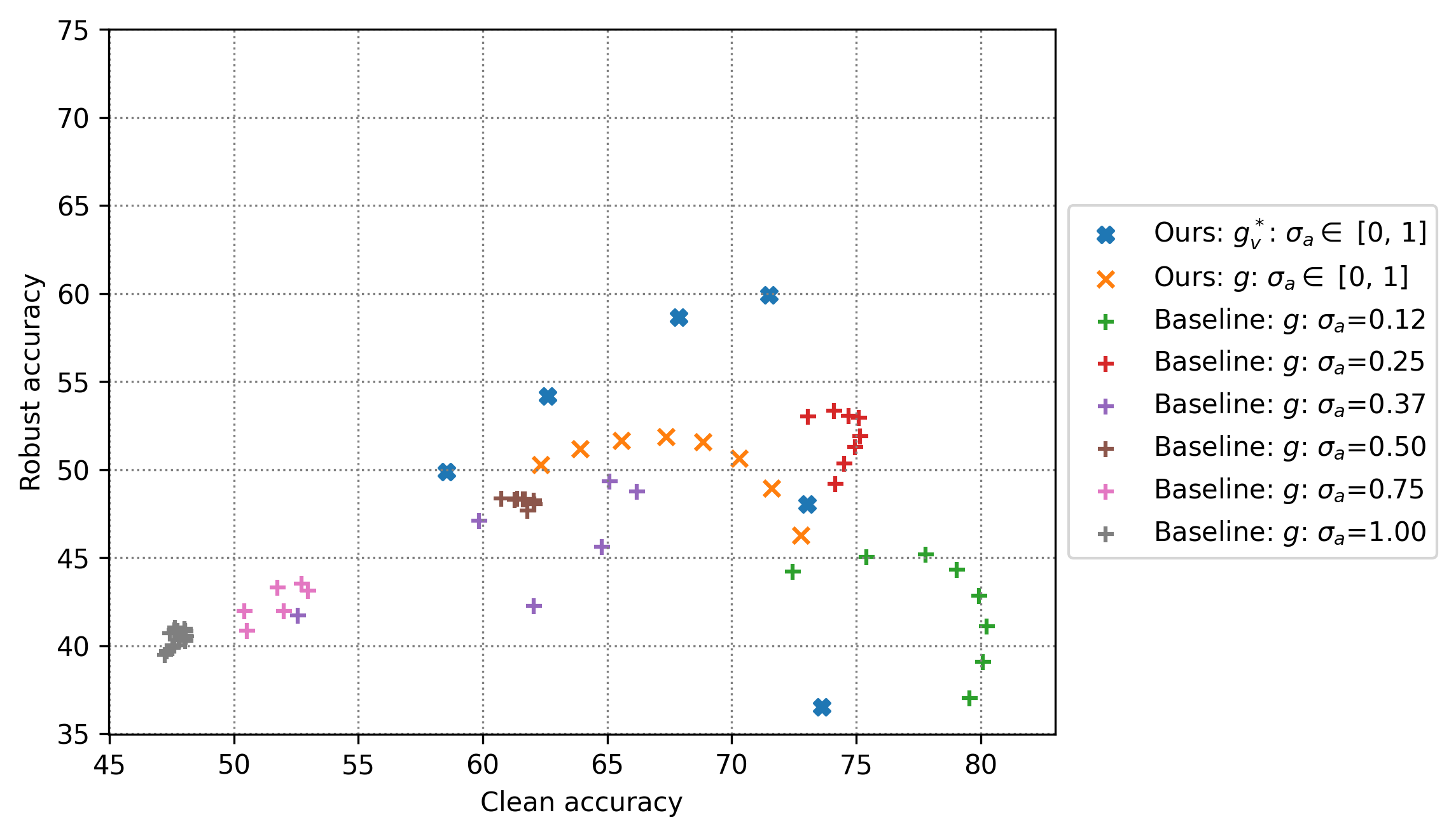}
        \caption{Perturbation budget for robust accuracy is $\gamma=0.5$.}
        \label{fig_appendix_experimental_bounds_examples_sigma0to1.00_gamma0.5}
  \end{subfigure}
  \caption{Clean and robust accuracy: fixed $\sigma_a$ training versus universal $\sigma_a$ training ($\sigma_a'=1.00$).}
\label{fig_appendix_universal_sigma_training_sigma1.00}
\end{figure}

Fig.~\ref{fig_appendix_universal_sigma_training_sigma0.25} shows results for $\sigma_a'=0.25$. 
Our method performed well against the baseline with $\sigma_a=0.12$, which was chosen since it is the mean of the range $[0,0.25]$ used in universal $\sigma_a$ training.
Although the performance of our method and the baseline with $\sigma_a=0.12$ both dropped significantly with $\gamma=0.5$, our method demonstrated better robustness.
Similar characteristics can be observed in Fig.~\ref{fig_appendix_universal_sigma_training_sigma0.50} and Fig.~\ref{fig_appendix_universal_sigma_training_sigma1.00}. 
In Fig.~\ref{fig_appendix_universal_sigma_training_sigma0.50}, our method with $\sigma_a'=0.50$ performed better than baseline with $\sigma_a=0.25$. Fig.~\ref{fig_appendix_universal_sigma_training_sigma1.00} indicates our method with $\sigma_a'=1.00$ performed better than baseline with $\sigma_a=0.50$.

\clearpage

\section{Selector training scheme}\label{appendix_model_training_scheme}
\begin{figure}[h]
  \centering
  \begin{subfigure}{0.5\linewidth}
\includegraphics[width=\columnwidth]{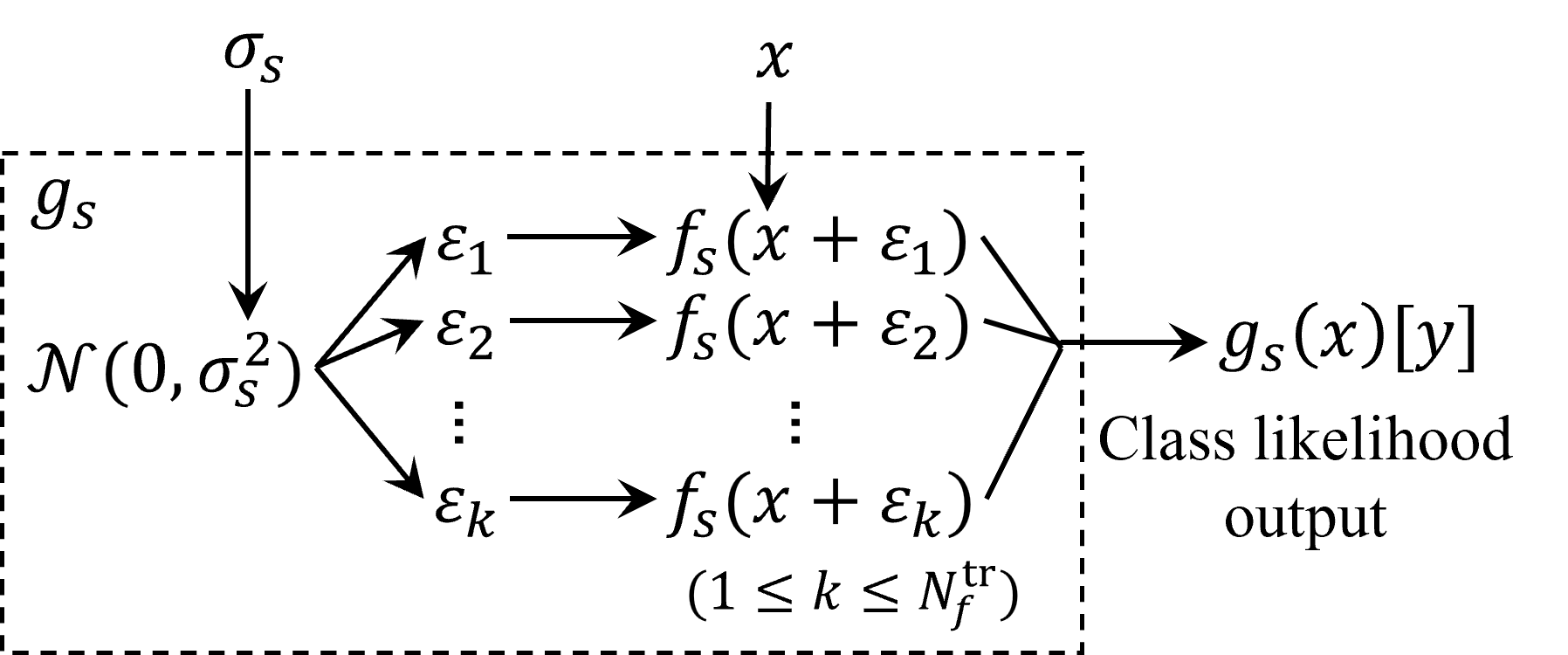}
        \caption{Process of soft smoothed classifier $g_s$.}
        \label{fig_appendix_training_scheme_soft_smoothed_classifier}
  \end{subfigure}\\
  \begin{subfigure}{0.55\linewidth}    
    \includegraphics[width=\columnwidth]{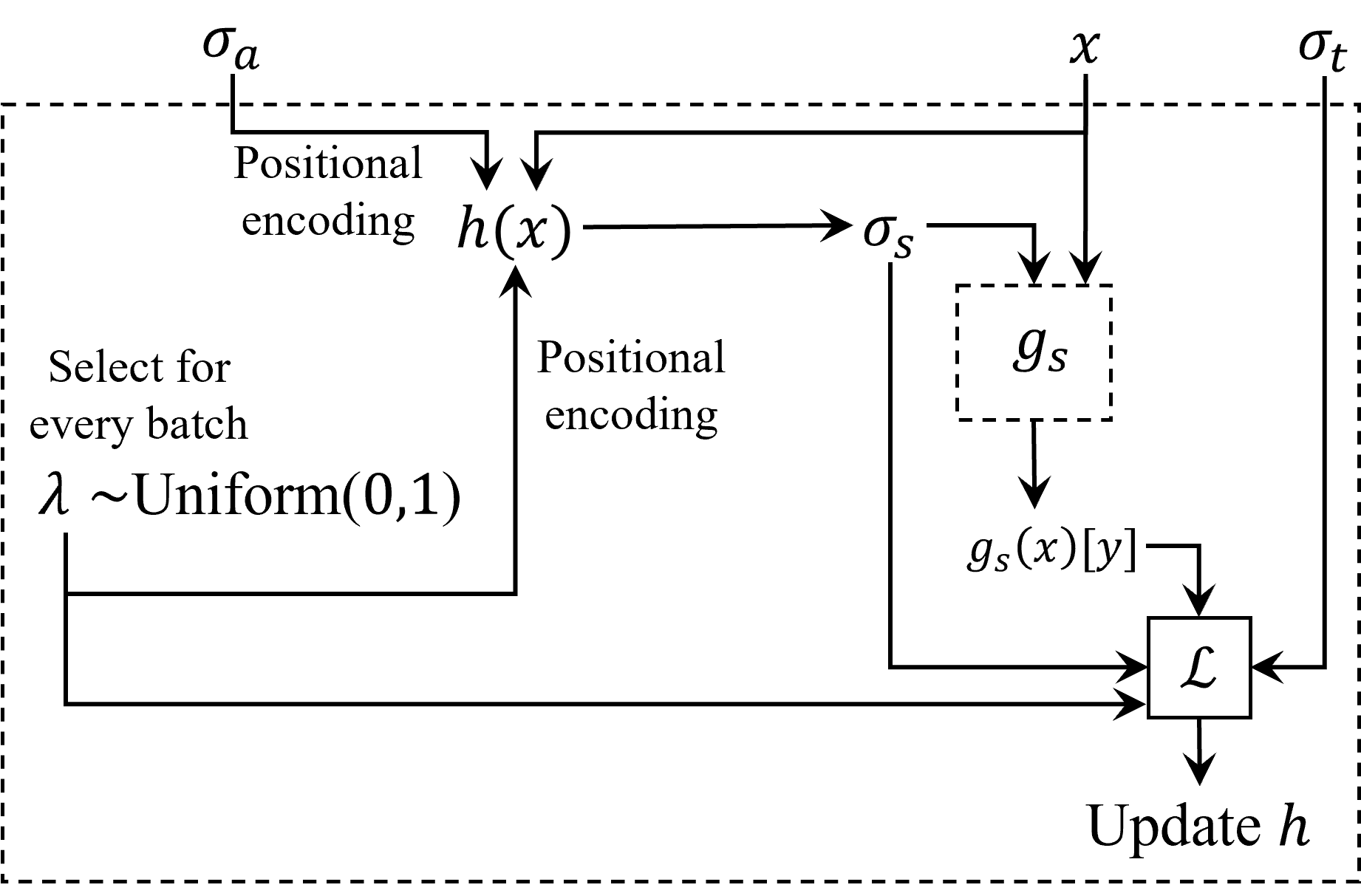}
        \caption{Training scheme for $g_v$.}
        \label{fig_appendix_training_scheme_g_v}
  \end{subfigure}\\
  \begin{subfigure}{0.55\linewidth}
\includegraphics[width=\columnwidth]{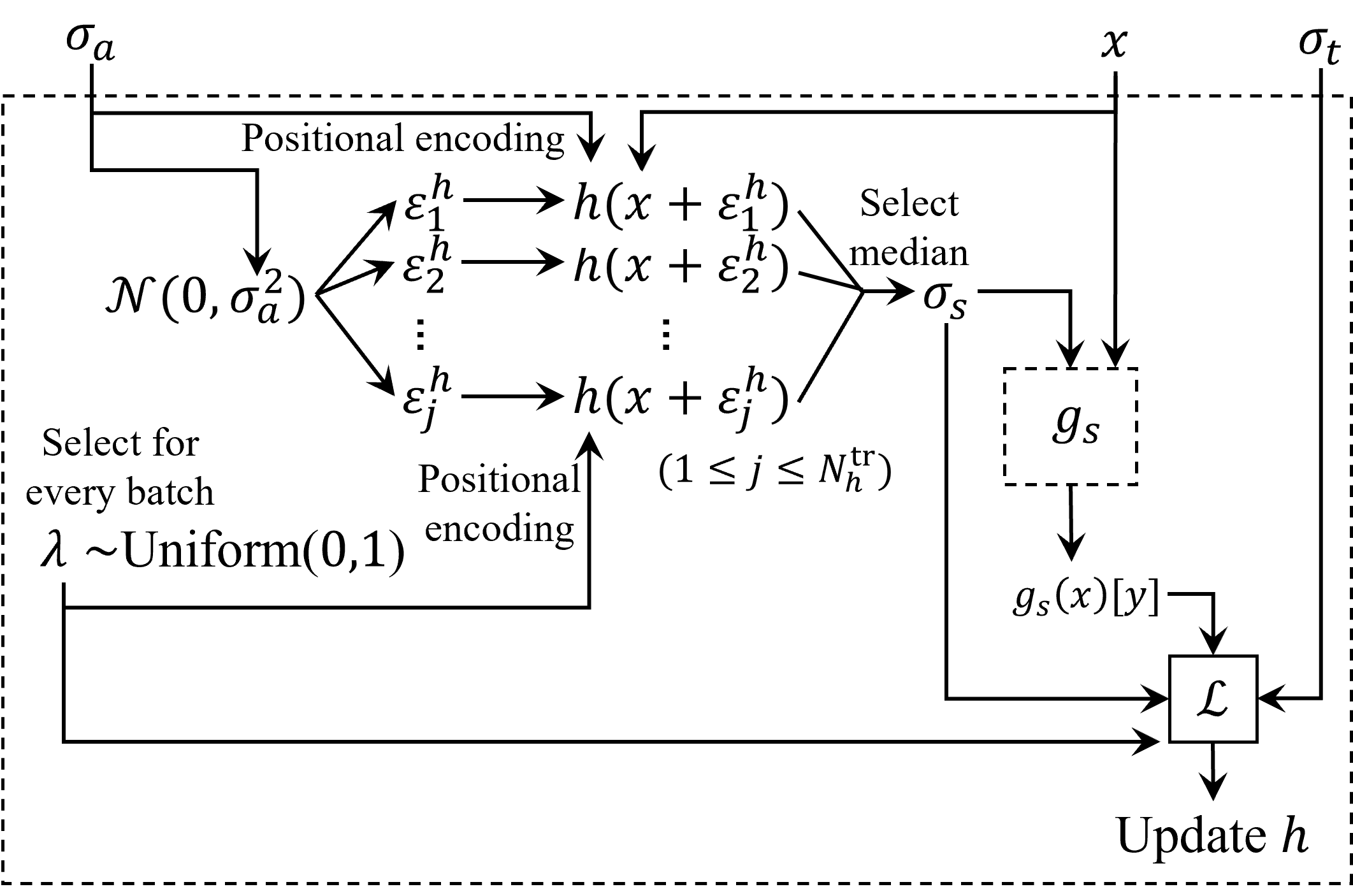}
        \caption{Training scheme for $g_v^*$.}
        \label{fig_appendix_training_scheme_g_v_star}
  \end{subfigure}
  \caption{Illustrations about training scheme for $g_v$ and $g_v^*$.}
\label{fig_appendix_selector_training_scheme}
\end{figure}

Fig.~\ref{fig_appendix_selector_training_scheme} depicts the training scheme of selector $h$ described in Algorithm~\ref{alg_training_process}. The training scheme utilizes soft smoothed classifier $g_s$ defined by~\eqref{eq_smoother_definition} as a component. The overview of $g_s$ is shown in Fig.~\ref{fig_appendix_training_scheme_soft_smoothed_classifier}. $N_f^\mathrm{tr}$ was set to 10 for all experiments presented in this paper. 
The loss function $\mathcal{L}$ to update $h$ is defined as~\eqref{eq_two_loss_functions_using_kl_divergence}. Fig.~\ref{fig_appendix_training_scheme_g_v} and Fig.~\ref{fig_appendix_training_scheme_g_v_star} show training scheme for $g_v$ and $g_v^*$, respectively. Fig.~\ref{fig_appendix_training_scheme_g_v_star} illustrates the process to generate $N_h^\mathrm{tr}$ samples of $\sigma_s$ and select median of the samples to simulate the process of median smoothing. We used $N_h^\mathrm{tr}= 10$ for all experiments in this paper.

\end{document}